\documentclass[lettersize,journal]{IEEEtran}
\usepackage{amsmath,amsfonts}
\usepackage{algorithmic}
\usepackage{algorithm}
\usepackage{array}
\usepackage[caption=false,font=normalsize,labelfont=sf,textfont=sf]{subfig}
\usepackage{textcomp}
\usepackage{stfloats}
\usepackage{url}
\usepackage{verbatim}
\usepackage{graphicx}
\usepackage{cite}

\usepackage{multirow}
\usepackage[table]{xcolor}
\usepackage[mathscr]{euscript}
\usepackage{hyperref}
\hypersetup{hypertex=true,
            colorlinks=true,
            linkcolor=blue,
            anchorcolor=blue,
            citecolor=blue}
            
\renewcommand{\textcolor}[2]{#2}
            
\begin{document}
\bibliographystyle{IEEEtran}

\title{Scale Propagation Network for Generalizable Depth Completion}

\author{Haotian Wang, Meng Yang,~\IEEEmembership{Member,~IEEE,} 
Xinhu Zheng,~\IEEEmembership{Member,~IEEE,} and Gang Hua,~\IEEEmembership{Fellow,~IEEE} 
\thanks{Manuscript received 15 March, 2024; revised 18 October, 2024. This work was supported by the National Science Foundation of China under Grants 62088102 and 62373298.}
\thanks{\textcolor{red}{Codes and models are available at at \href{https://github.com/Wang-xjtu/SPNet}{https://github.com/Wang-xjtu/SPNet}.}}
\thanks{H. Wang and M. Yang are with the Institute of Artificial Intelligence and Robotics, Xi'an Jiaotong University, Xi'an, P.R.China. (e-mails: wht\_sxchina@stu.xjtu.edu.cn, mengyang@mail.xjtu.edu.cn)

X. Zheng is with the Intelligent Transportation Thrust of the Systems Hub, Hong Kong University of Science and Technology (GZ), Guangzhou, P.R.China. (e-mail: xinhuzheng@hkust-gz.edu.cn)

Gang Hua is with the Multimodal Experiences Lab, Dolby Laboratories Inc, Los Angeles, CA, USA, and 
affiliated with the Institute of Artificial Intelligence and Robotics, Xi'an Jiaotong University, Xi'an 710049, P.R.China. (e-mail: ganghua@gmail.com)
}
}

\markboth{IEEE Transactions on Pattern Analysis and Machine Intelligence,~Vol.~x, No.~x, x~2024}%
{Shell \MakeLowercase{\textit{et al.}}: A Sample Article Using IEEEtran.cls for IEEE Journals}

\IEEEpubid{0000--0000/00\$00.00~\copyright~2021 IEEE}

\maketitle

\begin{abstract}
Depth completion, inferring dense depth maps from sparse measurements, is crucial for robust 3D perception. Although deep learning based methods have made tremendous progress in this problem, these models cannot generalize well across different scenes that are unobserved in training, posing a fundamental limitation that yet to be overcome. A careful analysis of existing deep neural network architectures for depth completion, which are largely borrowing from successful backbones for image analysis tasks, reveals that a key design bottleneck actually resides in the conventional normalization layers. These normalization layers are designed, on one hand, to make training more stable, on the other hand, to build more visual invariance across scene scales. However, in depth completion, the scale is actually what we want to robustly estimate in order to better generalize to unseen scenes. To mitigate, we propose a novel scale propagation normalization (SP-Norm) method to propagate scales from input to output, and simultaneously preserve the normalization operator for easy convergence. More specifically, we rescale the input using learned features of a single-layer perceptron from the normalized input, rather than directly normalizing the input as conventional normalization layers. We then develop a new network architecture based on SP-Norm and the ConvNeXt V2 backbone. We explore the composition of various basic blocks and architectures to achieve superior performance and efficient inference for generalizable depth completion. Extensive experiments are conducted on six unseen datasets with various types of sparse depth maps, i.e., randomly sampled 0.1\%/1\%/10\% valid pixels, 4/8/16/32/64-line LiDAR points, and holes from Structured-Light. Our model consistently achieves the best accuracy with faster speed and lower memory when compared to state-of-the-art methods.
\end{abstract}


\section{Introduction}
\IEEEPARstart{D}{epth} data is crucial for 3D perception \cite{wang2022depth} \cite{wang2023rgb} in widely downstream applications such as SLAM \cite{campos2021orb}, 3D object detection \cite{wang2019pseudo}, 3D odometry \cite{zhang2014real}, and 3D reconstruction \cite{chang2017matterport3d}. A popular approach in both academia and industry is to acquire sparse depth maps by depth sensors such as 4/8/16/32/64-line LiDAR, Structured-Light, and Time-of-Flight \cite{park2020non} \cite{imran2021depth} \cite{zhang2023completionformer}. The task of depth completion aims to infer dense depth maps from sparse depth maps and visual images. In recent years, deep learning based methods have made tremendous progress for this task through various techniques such as semantic cues \cite{nazir2022semattnet} \cite{zhang2021multitask}, surface normal \cite{zhang2018deep} \cite{qiu2019deeplidar}, spatial propagation networks \cite{park2020non} \cite{cheng2019learning}, and advanced backbones \cite{zhang2023completionformer} \cite{ma2019self} \cite{shao2022towards}.

However, these deep learning based models still focus on a single scene by training and testing on NYUv2 \cite{silberman2012indoor} or KITTI \cite{geiger2012we}, which cannot generalize well across different scenes that are unobserved in training. This paper aims to explore the generalization issue of depth completion across different scenes, namely \textit{generalizable depth completion}.

Current deep neural network architectures of depth completion are largely borrowing from successful backbones for image analysis tasks such as ResNet \cite{he2016deep}, UNet \cite{ronneberger2015u}, and Vision Transformers (ViTs) \cite{dosovitskiy2020image}. These network architectures were originally designed to build more visual invariance across different scales of scenes. For example, in image classification or semantic segmentation, the focus of these networks is to robustly predict target categories by finding relatively larger values in the normalized probabilities via a Softmax function, even though the scale changes drastically across scenes.

However, in depth completion, it requires inferring absolute depth values between scenes and camera plane, which additionally involves specific scales of the scenes \cite{ranftl2020towards}. 

The invariance of scene scales in existing network architectures can result in inaccurate scaling between inferred depth values and real distances. This issue can limit the generalization of depth completion in unseen scenes. 

Current models of depth completion work well when training and testing on a single dataset \cite{ma2019self} \cite{tang2020learning} \cite{nazir2022semattnet}, because scales of test scenes can be well learned in training. Unfortunately, scales of test scenes are generally unknown in training for generalizable depth completion \cite{ranftl2020towards}.

\IEEEpubidadjcol

We make a careful analysis of current network architectures for depth completion. We observe that a key design bottleneck of these networks resides in the conventional normalization layers such as Batch Normalization (BN) \cite{ioffe2015batch}, Instance Normalization (IN) \cite{ulyanov2016instance}, and Layer Normalization (LN) \cite{ba2016layer}.

These normalization layers have been widely used as basic components in modern networks. They facilitate the convergence of deep neural networks by normalizing scales of input data to unit scales along the dimensions of batch, spatial, and channel, respectively. However, scales of input data cannot be easily restored in the output from normalized scales of input data, especially for unseen scenes. 

In generalizable depth completion, the scale is actually what we want to robustly estimate in the output in order to better generalize to unseen scenes. Because single visual images are scale-ambiguous \cite{xian2020structure} \cite{zhang2023self}, scales of inferred depth maps in the output are mainly estimated from sparse depth maps in the input. However, our careful analysis reveals that these conventional normalization layers hinder the propagation of scales from input to output (see Section \ref{section: sp-norm}).

One optional solution for this issue is non-normalization techniques such as ReZero \cite{bachlechner2021rezero}, SkipInit \cite{de2020batch}, and Fixup \cite{zhang2019fixup}. However, the lack of normalization layers can have a detrimental impact on the stability of deep neural networks \cite{touvron2021going}, especially for large models \cite{dosovitskiy2020image} \cite{liu2022convnet}.

\begin{figure*}[!t]
\centering
\includegraphics[width=\textwidth]{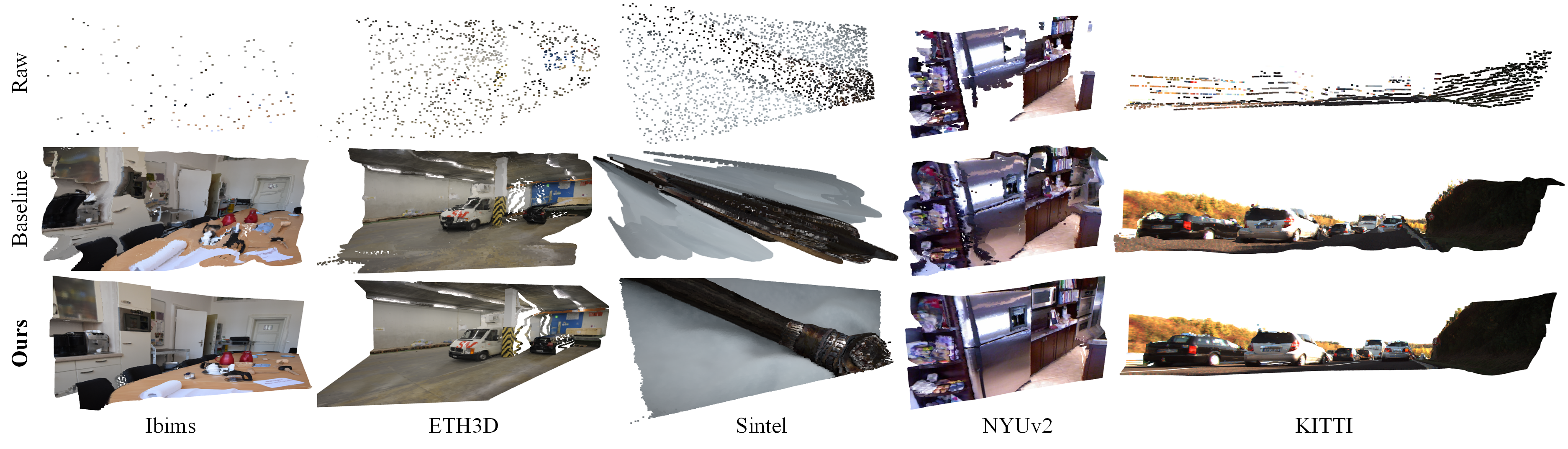}
\caption{Examples of generalizable depth completion across different scenes by our model and a recent SOTA baseline \cite{zhang2023completionformer}. Our model always infers accurate depth values and thereby well maintains the structure of objects in 3D view. In addition, our model has faster speed (126.6 vs 11.1 image/s) on a 3090 GPU.} 
\label{fig: examples}
\end{figure*}

In this paper, we first propose a novel scale propagation normalization method, namely SP-Norm, to overcome the limitations of conventional normalization layers and non-normalization techniques for generalizable depth completion. SP-Norm comprises a normalization operator, a Single-layer Perceptron (SLP), and a multiplier. More specifically, it is implemented by rescaling the input using learned features of the SLP from the normalized input, rather than directly normalizing the input as conventional normalization layers. 

Our analysis manifests that SP-Norm well enables the propagation of scales from input to output (see Section \ref{section: sp-norm}). In addition, it simultaneously preserves the normalization operator for easy convergence of deep neural networks. Therefore, it ensures the generalization ability of deep neural networks for depth completion across different scenes.  

We then develop a new network architecture for generalizable depth completion based on our SP-Norm and the ConvNeXt V2 backbone \cite{woo2023convnext}. ConvNeXt V2 successfully builds a paradigm of deep neural networks that leveraging large kernel convolutions. We explore the composition of different basic blocks and architectures to achieve superior performance and efficient inference in our task.

On one hand, we make several modifications to the basic block of ConvNeXt V2. Firstly, we replace all LN with our SP-Norm to enable the propagation of scales from input to output. Secondly, we remove the core operator, i.e., Global Response Normalization (GRN), of ConvNext V2 from our basic block. We find that GRN is harmful to depth completion possibly because features of sparse depth maps are suppressed by its reweighting strategy. Thirdly, we replace the activation function GELU with RELU to reduce the cost of memory and time. On the other hand, our network architecture comprises a heavyweight encoder and a lightweight decoder similar to \cite{kirillov2023segment}. The heavyweight encoder can provide large receptive fields and long-range relationships, while the lightweight decoder can accelerate inference. 

Our network is trained on a mixture of four datasets i.e., Matterport3D \cite{chang2017matterport3d}, HRWSI \cite{xian2020structure}, vKITTI \cite{gaidon2016virtual}, and UnrealCV \cite{wang2023g2}, and tested on six unseen datasets, i.e., Ibims \cite{koch2018evaluation}, KITTI \cite{geiger2012we}, NYUv2 \cite{silberman2012indoor}, DIODE \cite{vasiljevic2019diode}, ETH3D \cite{schops2017multi}, and Sintel \cite{butler2012naturalistic}. Extensive experiments are conducted on various types of sparse depth maps with randomly sampled 0.1\%/1\%/10\% valid pixels, 4/8/16/32/64-line LiDAR points, and holes from the Structured-Light. 

Our network consistently achieves the best accuracy with faster speed, lower FLOPs, and lower memory, when compared to recent state-of-the-art (SOTA) baselines with both officially released models by the authors and retrained models on our training data. Fig.\ref{fig: examples} shows several examples of our model and a recent SOTA baseline \cite{zhang2023completionformer}.

Our main contributions are summarized as follows:
\begin{enumerate}
\item{We analyze that conventional normalization layers limit the generalization of depth completion across scenes. We propose a novel SP-Norm to well propagate scene scales from input to output, and simultaneously preserve the normalization operator for easy convergence.}
\item{We develop a new network for generalizable depth completion based on SP-Norm and ConvNeXt V2. We explore the composition of basic blocks and architectures to achieve better performance and faster inference.}
\item{Extensive experiments on six unseen datasets with various types of sparse depth maps verify that our network consistently achieves the best accuracy with faster speed and lower memory compared to SOTA baselines.}
\end{enumerate}

\section{Related work}
\subsection{Depth completion}
Deep learning based methods have made tremendous progress for depth completion in recent years. These methods often introduce various techniques into deep neural networks. 
Semantic cues were introduced to help models well understand the composition of scenes \cite{nazir2022semattnet} \cite{zhang2021multitask}. Surface normal cues were incorporated into loss functions or network design to constrain structures of completed depth maps in 3D space \cite{zhang2018deep} \cite{qiu2019deeplidar} \cite{xu2019depth}. Some models well integrated features in 2D and 3D spaces using continuous convolutions \cite{chen2019learning}, graph propagation \cite{zhao2021adaptive}, and Bird's-eye view (BEV) representation \cite{zhou2023bev}. 

The most popular network architectures for depth completion are the spatial propagation networks (SPNs) \cite{park2020non} \cite{cheng2019learning} \cite{lin2022dynamic}. The SPNs can iteratively refine initial predictions of completed depth maps by learning an affinity matrix. However, this refinement generally requires more time for multiple iterations. The model \cite{wang2023lrru} reduced the iterative times from more than ten to four corresponding to four resolution stages. In addition, many models benefited from successful backbones of image analysis tasks, such as ResNet \cite{ma2019self}, UNet \cite{wang2023lrru}, and ViTs \cite{zhang2023completionformer} \cite{shao2022towards}. Some complicated architectures \cite{tang2020learning} \cite{zhang2020multiscale} \cite{chen2020laplacian} were also developed to fuse features from images and depth maps better. 

Current deep learning based methods have achieved significant success for depth completion on a single scene such as NYUv2 \cite{silberman2012indoor} or KITTI \cite{geiger2012we}. Nonetheless, their models could not generalize well across different scenes that are unobserved in training \cite{hu2022deep} \cite{wang2023g2}. This paper aims to address this generalization issue of depth completion.

\subsection{Network architecture}
Current network architectures of depth completion are largely borrowing from successful backbones of image analysis tasks, such as image classification \cite{he2016deep} and semantic segmentation \cite{zhao2017pyramid}.  
Inceptions \cite{szegedy2017inception}, ResNeXt \cite{xie2017aggregated}, MobileNets \cite{howard2019searching}, and EfficentNets \cite{tan2019efficientnet} incorporated group or depth-wise convolutions to convolution neural networks for better performance with fewer parameters. Recently, ViTs \cite{dosovitskiy2020image} provided powerful Transformer architectures for image analysis tasks. Swin Transformers \cite{liu2021swin} adopted self-attention in local window to accelerate inference and reduce memory. ConvNeXts \cite{liu2022convnet} \cite{woo2023convnext} explored powerful networks by applying the modernized design of ViTs to convolution neural networks. RepLKNet \cite{ding2022scaling} and InternImage \cite{wang2023internimage} respectively explored the advantages of large-kernel and deformable convolutions. 

These modern network architectures generally include three basic components: linear layers (e.g., convolution, fully connected layer, and SLP), activation functions (e.g., Sigmoid, ReLU, and GELU), and normalization layers (e.g., BN \cite{ioffe2015batch}, IN \cite{ulyanov2016instance}, and LN \cite{ba2016layer}). Some researchers attempted to remove these normalization layers by a learnable scaler with different initialization such as ReZero \cite{bachlechner2021rezero}, SkipInit \cite{de2020batch}, Fixup \cite{zhang2019fixup}, or by carefully designing basic blocks such as NF-ResNet \cite{brock2021characterizing} and NF-Net \cite{brock2021high}. However, LayerScale \cite{touvron2021going} reported that adopting normalization can facilitate the convergence of networks, especially in large models. 

This paper analyzes that a key design bottleneck of current network architectures actually resides in the conventional normalization layers for depth completion across different scenes. Therefore, we propose a novel SP-Norm method as well as a new network architecture based on the ConvNeXt V2 for generalizable depth completion.

\section{Scale propagation normalization}
\label{section: sp-norm}
\subsection{Scale propagation property}
The goal of generalizable depth completion is to infer a dense depth map $z$ from a sparse depth map $d$ with the guidance of a visual image $I$ in unseen scenes. The input $I$ and $d$ are acquired by cameras and depth sensors respectively, both of which vary in different scenes. 

\begin{figure}[!t]
\centering
\includegraphics[width=\columnwidth]{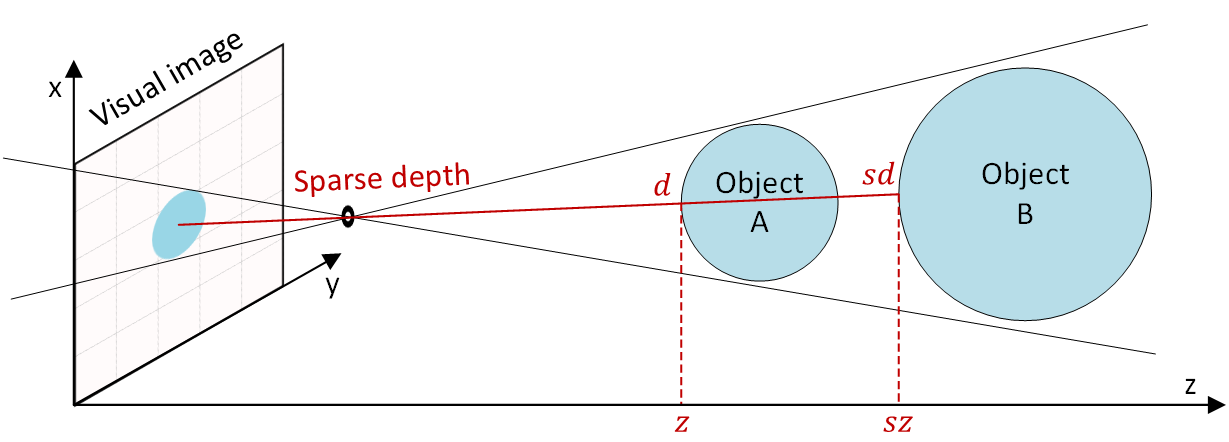}
\caption{\textcolor{red}{Illustration of the SP-property. The ambiguous scales of output depth values $z$ or $sz$ can be determined by input sparse points $d$ or $sd$, respectively.}} 
\label{fig: sp property}
\end{figure}

Because scene scales are ambiguous for visual image $I$ \cite{ranftl2020towards} \cite{zhang2023self} in the input, scales of completed depth maps $z$ in the output are mainly determined by sparse depth maps $d$ in the input. \textcolor{red}{Therefore, it requires the scales of input $d$ and output $z$ to be always consistent with each other, as illustrated in Fig. \ref{fig: sp property}.} That is, when the input $d$ is scaled to $sd$ by a scale factor $s$, the output $z$ should be proportionally scaled to $sz$. This fundamental property is referred to \textit{scale propagation} in this paper, which is represented as $z \propto d$ for simplicity. 

We take the mean and variance to approximately examine the relationship between the two variables $z$ and $d$. It can be easily derived that the mean and variance of \textcolor{red}{the input $d$ should be proportional to the ones of the output $z$,} respectively. That is, $E(z) \propto E(d)$ and $D(z) \propto D(d)$, where $E(.)$ and $D(.)$ are functions of mean and variance. This property is concluded as follows.

\textit{\textbf{SP-property}. In generalizable depth completion, input sparse depth $d$ and output dense depth $z$ should always satisfy $z \propto d$. It can be examined by $E(z) \propto E(d)$ and $D(z) \propto D(d)$ approximately.}

Failure to satisfy this property can degrade the generalization ability of depth completion in unseen scenes. In the following, we examine conventional normalization layers and our SP-Norm with this property.

\begin{figure}[!t]
\centering
\includegraphics[width=0.9\columnwidth]{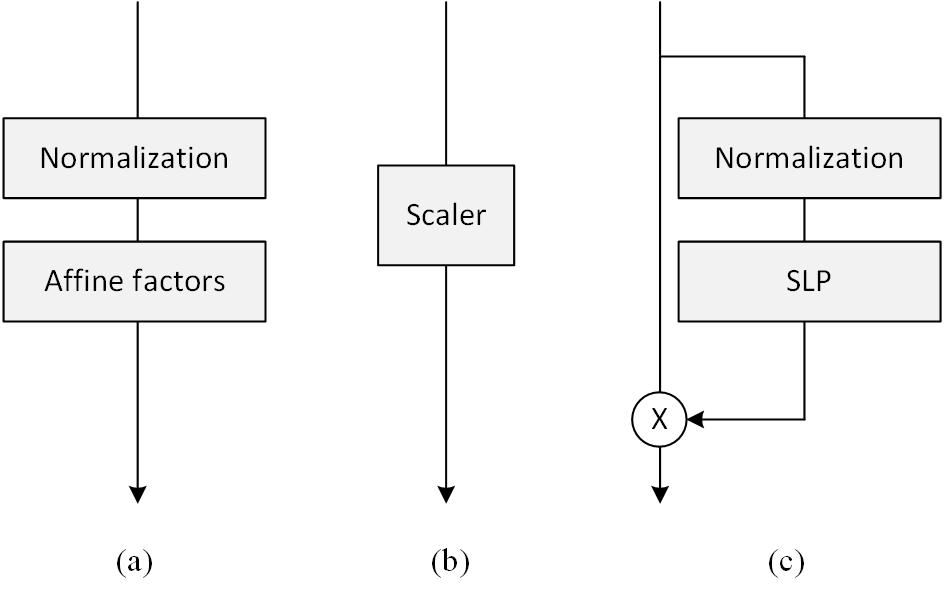}
\caption{Different normalization strategies. (a) conventional normalization layers (e.g., BN, IN, and LN), (b) non-normalization techniques (e.g., ReZero, SkipInit, and Fixup), and (c) our SP-Norm.}
\label{fig: norm_diff}
\end{figure}

\subsection{Conventional normalization}
The normalization layers have been a basic component in modern networks including BN \cite{ioffe2015batch}, IN \cite{ulyanov2016instance}, and LN \cite{ba2016layer}. These layers similarly comprise a normalization operator and affine factors in Fig. \ref{fig: norm_diff}(a), though they are respectively operated along the dimensions of batch, spatial, and channel.

\textcolor{red}{More specifically, the input of the normalization layer $d_i$ is first normalized by the mean $\bar{d}$ and the standard deviation $\delta_d$ in Eqn. (\ref{eq: norm})}, where $i$ denotes pixel location. Then, normalized data $\hat{d} _ i$ is rescaled to output data $z^{cd}_i$ using learnable affine factors $\alpha_i$ and $\beta_i$ in Eqn. (\ref{eq: affine}).
\begin{equation}
\label{eq: norm}
{\hat{d}_i} = ( d_i - \bar{d} ) / {\delta_d}.
\end{equation}
\begin{equation}
\label{eq: affine}
z^{cd}_i = {\alpha_i} {\hat{d}_i} + {\beta_i}.
\end{equation}

\subsubsection{SP-Property in conventional normalization}
We examine the SP-property in conventional normalization for generalizable depth completion. By taking the mean and variance of Eqn. (\ref{eq: affine}), we have
\begin{equation}
\label{eq: spfortrdnorm}
\left\{
\begin{aligned}
    & E({z^{cd}_i}) = E({\beta_i}), \\
    & D({z^{cd}_i}) = D({\alpha_i}) + E({\alpha_i})^2 + D({\beta_i}).
\end{aligned}
\right.
\end{equation}
The detailed derivation of Eqn. (\ref{eq: spfortrdnorm}) can be found in Appendix. 

It is seen that the mean $E({z^{cd}_i})$ and the variance $D({z^{cd}_i})$ of the output data $z^{cd}_i$ are determined by the affine factors $\alpha_i$ and $\beta_i$, both of which are constant during testing. However, the mean $E(d_i)$ and the variance $D(d_i)$ of the input data $d_i$ can vary in different scenes.  It is also impractical to learn $E(d_i)$ and $D(d_i)$ of unseen scenes by affine factors $\alpha_i$ and $\beta_i$ in training, because testing data may be totally different from training data in unseen scenes. In conclusion, the SP-property cannot be always satisfied in conventional normalization layers. It can limit the model generalization of depth completion across different scenes. 

\subsubsection{Initial state of conventional normalization} We additionally analyze the initial state of conventional normalization layers based on Eqn. (\ref{eq: spfortrdnorm}). The initial states of affine factors are $E(\alpha_i)=1$, $D(\alpha_i)=0$, $E(\beta_i)=0$, and $D(\beta_i)=0$. Therefore, we can get a specific initial state of Eqn. (\ref{eq: spfortrdnorm}), i.e., $E({z^{cd}_i})=0$ and $D({z^{cd}_i})=1$. 

\textcolor{red}{To analyze the specific initial state of the input $d_i$, we suppose that the network is initialized by Xavier Normal \cite{glorot2010understanding} and its first layer is a convolution layer. Therefore, $d_i$ is the output of the convolution layer with an initial state as}
\begin{equation}
\label{eq: initinput}
\left\{
\begin{aligned}
    & E(d_i) = 0, \\
    & D(d_i) = 2{n^0}/({n^0}+{n^1})(D(d^0_i)+E(d^0_i)^2),
\end{aligned}
\right.
\end{equation}
where ${n^0}$ and ${n^1}$ are the input dimension and output dimension of the convolution layer, respectively. $E(d^0_i)$ and $D(d^0_i)$ are the mean and variance of input data for this convolution layer. The detailed derivation of Eqn. (\ref{eq: initinput}) can be found in Appendix. 

It is clear that the SP-property $D(z^{cd}_i) \propto D(d_i)$ is not satisfied, because $D({z^{cd}_i})$ always equals to 1 while $D(d_i)$ will vary according to the input data $d^0_i$ of the convolution layer. It indicates that the conventional normalization layers do not satisfy the SP-property at the beginning of the training. 

One optional solution is adopting non-normalization techniques such as ReZero \cite{bachlechner2021rezero}, SkipInit \cite{de2020batch}, and Fixup \cite{zhang2019fixup}. In these layers, input data is directly rescaled using a learnable scalar without the normalization operator as Fig. \ref{fig: norm_diff}(b). However, normalization has been a crucial component in modern networks. Its removal has a detrimental impact on the stability of the network \cite{touvron2021going}, especially in large models \cite{dosovitskiy2020image} \cite{liu2022convnet}. We provide the ablation studies of different normalization strategies in Section \ref{section:ablation}.

\begin{figure*}[!t]
\centering
\includegraphics[width=0.95\textwidth]{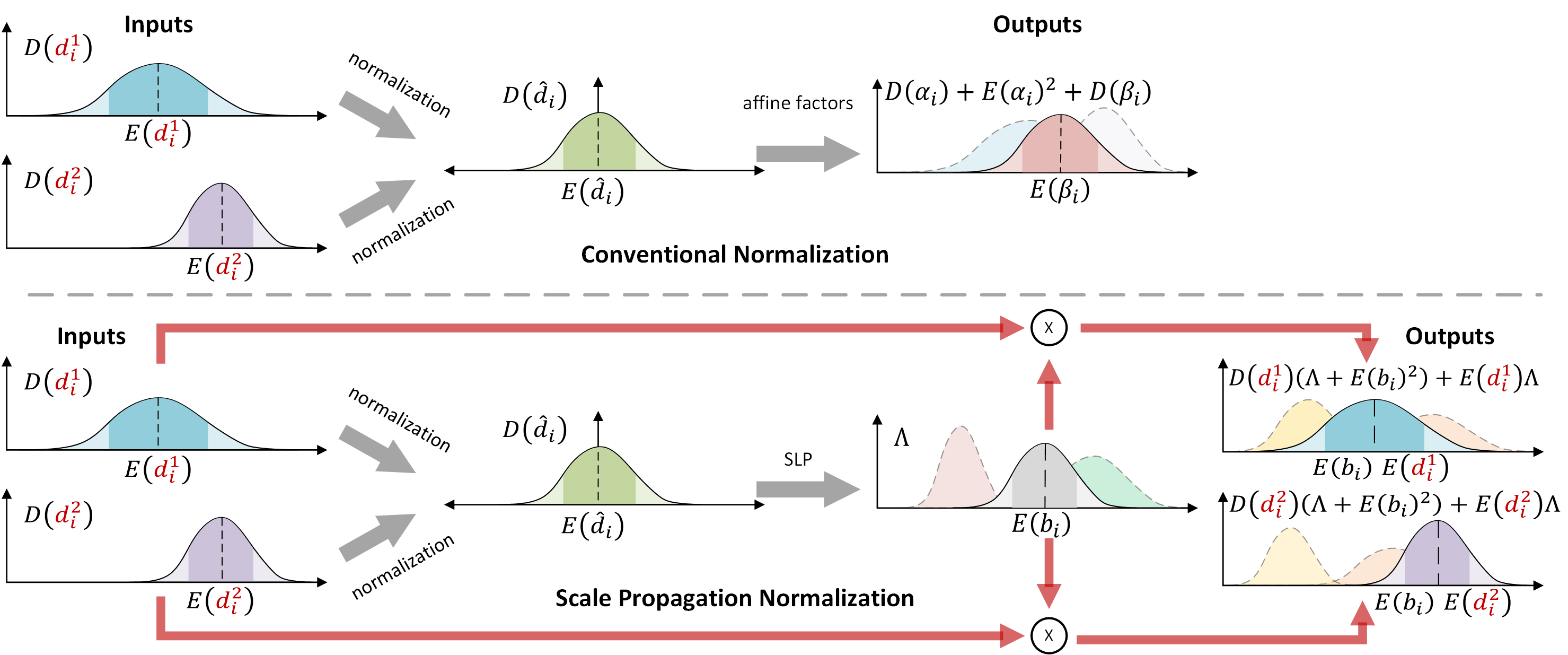}
\caption{\textcolor{red}{Illustration of conventional normalization and our SP-Norm. In conventional normalization, multiple inputs $d^1_i$ and $d^2_i$ with different scales are mapped to one output with the same scale. The output scale solely depends on learnable affine factors $\alpha_i, \beta_i$. In contrast, our SP-Norm can preserve the varying scales of the inputs to the outputs. The output scales jointly depend on both the scales of the inputs $d^1_i, d^2_i$ and learnable parameters $\omega_{ij}, b_i$ of SLP.}} 
\label{fig: illustrationNorm}
\end{figure*}

\subsection{SP-Norm}
We develop a novel scale propagation normalization, namely SP-Norm, to address the limitations of conventional normalization layers and non-normalization techniques in generalizable depth completion. Our SP-Norm is illustrated in Fig. \ref{fig: norm_diff}(c). \textcolor{red}{For efficiency and simplicity}, it comprises three components: a normalization operator, an SLP, and a multiplier. It is implemented by rescaling input using learned features of SLP from normalized input, rather than directly normalizing input as conventional normalization layers. 

\textcolor{red}{More specifically, the input of the normalization layer $d_i$ is first normalized to $\hat{d}_i$ following Eqn. (\ref{eq: norm}).} Normalized data $\hat{d}_i$ is then fed into the SLP to generate learned features. After that, input data $d_i$ is rescaled to output data $z^{sp}_i$ using learned features of the SLP. Our SP-Norm can be expressed as
\begin{equation}
\label{eq: spnorm}
{z^{sp}_i} = ({\sum_{j=1}^n} w_{ij} {\hat{d}_j} + {b_i}) {d_i},
\end{equation}
where $w_{ij}$ and $b_i$ denote learnable parameters of the SLP, $n$ denotes channel number of input data, and $j$ also denotes pixel location. Notably, our SP-Norm is operated along channel dimension.

\subsubsection{SP-property in SP-Norm} We examine the SP-property of our SP-Norm. By taking the mean and variance of Eqn. (\ref{eq: spnorm}), we have
\begin{equation}
\label{eq: sppforspnorm}
\left\{
\begin{aligned}
    & E({z^{sp}_i}) = E(b_i) E(d_i), \\
    & D({z^{sp}_i}) = D(d_i) ({\Lambda} + {E(b_i)^2}) + {E(d_i)^2}{\Lambda },
\end{aligned}
\right.
\end{equation}
where ${\Lambda}=n(D(w_{ij}) + E(w_{ij})^2) + D(b_i)$ is only determined by learnable parameters $w_{ij}$ and $b_i$ of the SLP. The detailed derivation of Eqn. (\ref{eq: sppforspnorm}) can be found in Appendix. 

We then analyze the SP-property of our SP-Norm based on Eqn. (\ref{eq: sppforspnorm}) in two cases.

\textbf{Case 1}: $E(d_i)=0$. In this case, the mean of output data is $E({z^{sp}_i})=0$. It is always proportional to the mean of input data $E(d_i)$. The variance of the output data is $D({z^{sp}_i})=D(d_i)({\Lambda} + E(b_i)^2)$. It is also proportional to the variance of input data $D(d_i)$, because ${\Lambda}$ and $E(b_i)$ are constant during testing.

\textbf{Case 2}: ${E(d_i)} \neq 0$. In this case, the mean of output data is $E({z^{sp}_i})=E(b_i)E(d_i)$. It is always proportional to the mean of input data $E(d_i)$, because $E(b_i)$ is constant during testing. The variance of output data is $D({z^{sp}_i})=D(d_i){E(b_i)^2}$, when the learnable variable ${\Lambda}$ equals to 0. It can be always proportional to the variance of input data $D(d_i)$ as well, because ${\Lambda}$ can be automatically adjusted by learning on training data.

In conclusion, our SP-Norm can always learn to satisfy the \textit{SP-property} for generalizable depth completion. It thereby well propagates scales from input to output, and simultaneously preserves the normalization operator for easy convergence. These two characteristics respectively overcome the limitations of conventional normalization layers and non-normalization techniques. By comparison, we have analyzed in the previous subsection that the SP-property cannot be always satisfied in conventional normalization layers. It limits the application in generalizable depth completion. \textcolor{red}{The difference of conventional normalization layers and our SP-Norm is further illustrated in Fig. \ref{fig: illustrationNorm}.}

\begin{figure}[!t]
\centering
\includegraphics[width=0.9\columnwidth]{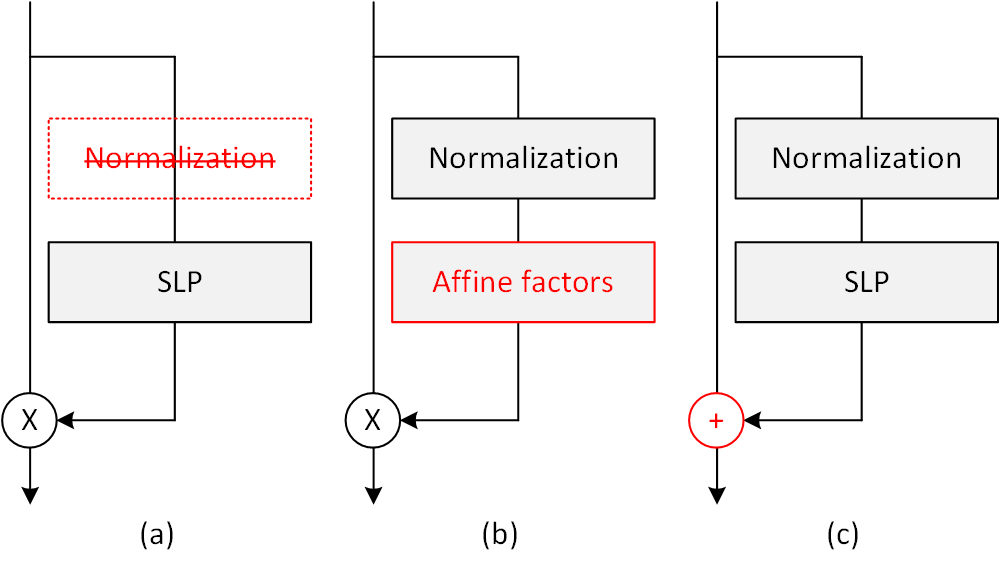}
\caption{Three variants of our SP-Norm. (a) removing the normalization operator, (b) replacing the SLP with the affine factors, and (c) replacing the multiplier with an adder.}
\label{fig: variants}
\end{figure}

\subsubsection{Initial state of SP-Norm} We also analyze the initial state of our SP-Norm based on Eqn. (\ref{eq: sppforspnorm}). Similarly, in a network initialized by Xavier Normal, we have $E(w_{ij})=0$, $D(w_{ij})={1/n}$, $E(b_i)=0$, and $D(b_i)=0$ for the SLP.  Therefore, we get a specific state of Eqn. (\ref{eq: sppforspnorm}) , i.e., $E({z^{sp}_i}) = 0$ and $D({z^{sp}_i}) =D(d_i)$. It is clear that the SP-property $E(z^{sp}_i) \propto E(d_i)$ and $D(z^{sp}_i) \propto D(d_i)$ are satisfied, because $E(d_i)=0$ based on Eqn. (\ref{eq: initinput}). It indicates that our SP-Norm satisfies the SP-property at the beginning of training.  

Furthermore, Eqn. (\ref{eq: initinput}) has shown that the mean of output data is zero for a convolution layer. Eqn. (\ref{eq: sppforspnorm}) has shown that the mean of output data is zero for our SP-Norm. The output data of the current layer is the input data of the next layer. Therefore, it indicates that the input data of next SP-Norm also has zero mean even stacking multiple layers of convolution or SP-Norm. This situation satisfies \textbf{Case 1} of Eqn. (\ref{eq: sppforspnorm}) for the SP-property.

\subsubsection{Role of SP-Norm components} We further explore the role of each component in our SP-Norm, i.e., the normalization operator, the SLP, and the multiplier. All these three components are crucial for our SP-Norm. On one hand, the normalization operator is used to improve the convergence of deep neural networks similar to conventional normalization layers. On the other hand, the SLP and the multiplier are jointly utilized to enable scale propagation from input to output to satisfy the SP-property. 

The role of the three components in our SP-Norm can be observed from three variants in Fig. \ref{fig: variants}. First, we remove the normalization operator in Fig. \ref{fig: variants}(a). This modification will lead to convergence failure of the network.
Second, we replace the SLP with affine factors of the conventional normalization layers in Fig. \ref{fig: variants}(b). This modified SP-Norm still satisfies the SP-property, however, it will lead to degraded performance. Third, we replace the multiplier with the widely used adder of residual learning in Fig. \ref{fig: variants}(c). This modification directly results in our SP-Norm not satisfying the SP-property, thus it will consequently lead to convergence failure of the network. The ablation studies of these three components in our SP-Norm are shown in Section \ref{section:ablation}.

\begin{figure*}[!t]
\centering
\includegraphics[width=\textwidth]{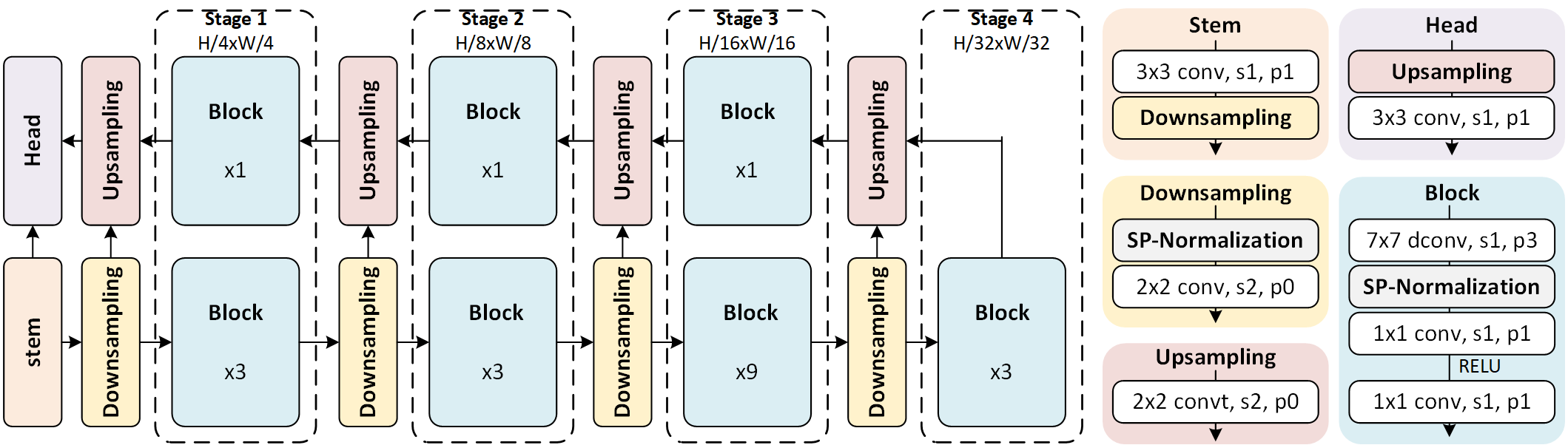}
\caption{The framework of our network. “conv”, “convt”, and “dconv” indicate convolution, transposed convolution, and depth-wise convolution, respectively. “s1” and “p1” indicate stride 1 and padding 1.}
\label{fig: network}
\end{figure*}

\section{Network Architecture}
We develop a new network architecture for generalizable depth completion based on our SP-Norm and the ConvNeXt V2  \cite{woo2023convnext}. ConvNeXt V2 is a successful paradigm of deep neural networks recently that leveraging large kernel convolutions. We expect that our network can benefit from this backbone.

Our network mainly comprises two parts including basic block and overall architecture in Fig. \ref{fig: network}. The goal of our network is to achieve superior performance and efficient inference for generalizable depth completion. Therefore, the whole network is designed on the basis of our SP-Norm.

\subsection{Basic block}
We explore the composition of our basic block for generalizable depth completion. The basic block of ConvNeXt V2 \cite{woo2023convnext} comprises LN, GELU, GRN, depth-wise convolution with large-kernel, and point-wise convolutions. 

Firstly, we fully replace LN with our SP-Norm to ensure the scale propagation of the network. The reason for this modification has been discussed in the last section. 

Secondly, we remove the GRN from our basic block. The GRN is the core operator of ConvNeXt V2, which is used to improve feature diversity. We find it is harmful to depth completion. The GRN is realized by reweighting features on channel dimension based on their global intensities of spatial dimension. However, our task additionally involves sparse depth maps in the input compared to image analysis tasks. There are generally a large quantity of invalid pixels with zero intensities in sparse depth maps. Therefore, features from sparse depth maps may be suppressed by GRN, leading to inaccurate depth prediction. Furthermore, removing the GRN is helpful to reduce the cost of memory and time. 

Thirdly, our basic block adopts the activation function RELU instead of the GELU. The reason is that ConvNeXt \cite{liu2022convnet} reports that the accuracy stays unchanged or degrades when replacing RELU with GELU, while the GELU requires more cost of memory and time. The composition of our basic block is confirmed in the ablation studies in Section \ref{section:ablation}.

\begin{table}[!t]
\caption{ Model configurations of “Tiny”, “Small”, “Base”, and “Large”. \label{tab:config}}
\centering
\resizebox{\columnwidth}{!}{
\begin{tabular}{lcccc}
\hline
      & Blocks     & Channels                 &  Params.   & Speed          \\
      & (4 stages) & (6 resolutions) & (M.) & (image/s)   \\
\hline
Tiny  & [3,3, 9,3]& [24,48, 96,192,384, 768]& 35.0       & 126.6           \\
Small & [3,3,27,3] & [24,48, 96,192,384, 768]& 59.3       & 77.6            \\
Base  & [3,3,27,3] & [32,64,128,256,512,1024] & 105.0      & 76.7            \\
Large & [3,3,27,3] & [48,96,192,384,768,1536] & 235.5      & 60.2            \\
\hline
\end{tabular}
}
\end{table}

\subsection{Architecture}
Our network architecture comprises a heavyweight encoder and a lightweight decoder similar to \cite{kirillov2023segment}. The heavyweight encoder can provide large receptive fields and long-range relationships, while the lightweight decoder can accelerate inference. 

The basic blocks are located at four stages of networks on the different resolutions {1/4, 1/8, 1/16, 1/32} for a feature pyramid. The heavyweight encoder contains more than eighteen basic blocks while the lightweight decoder only contains three basic blocks. The down-sampling layer comprises an SP-Norm and 2×2 stride convolution. The up-sampling layer comprises a single 2×2 transposed convolution without normalization and activation function for fast inference. A 3×3 convolution in the stem and a 3×3 convolution in the head are to generate accurate structures in the original resolution. We also adopt long-range skip connections to fuse features from the encoder and decoder in the resolution {1, 1/2, 1/4, 1/16, 1/32}. 

To verify the stability of our network, we provide four versions of our network including ``Tiny", ``Small", ``Base", and ``Large". The number of blocks and channels for these versions follows the settings of ConvNeXts. The detailed configurations of these four versions are shown in Table \ref{tab:config}. We verify these versions of our network in Section \ref{section:ablation}. Larger models can consistently achieve better performance, which indicates that our network can be stably trained even with more parameters. 

\begin{figure}[!th]
\centering
\includegraphics[width=0.9\columnwidth]{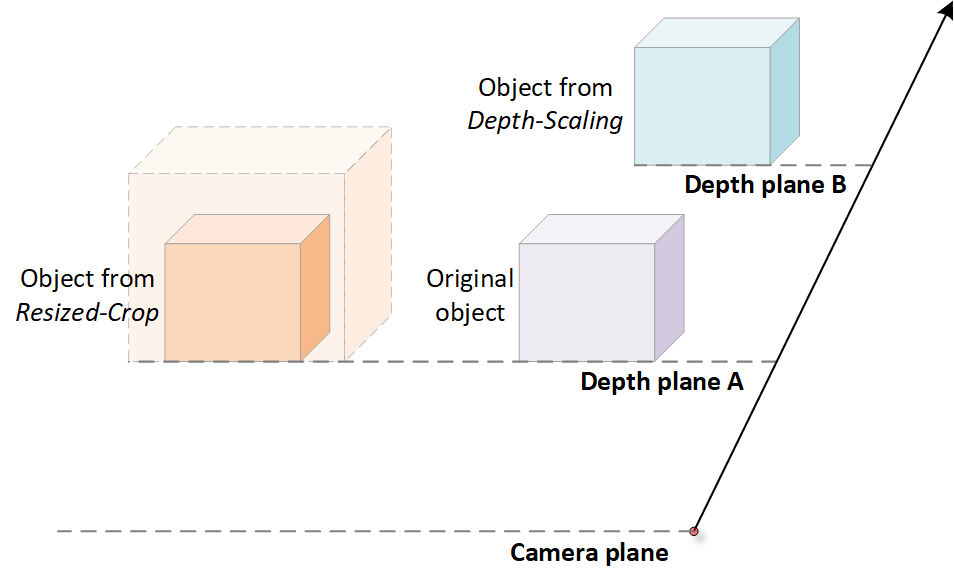}
\caption{Illustration of data augmentation with Resized-Crop and Depth-Scaling.}
\label{fig: augmentation}
\end{figure}

\subsection{Implementations}
\label{section: implementation}
\subsubsection{Data augmentation} Our network requires training data with diverse scene scales for generalizable depth completion. We adopt two strategies to simulate diverse scales on public datasets including Resized-Crop and Depth-Scaling. 

Resized-Crop randomly crops and resizes an image patch. It was originally used in image classification \cite{liu2021swin} and semantic segmentation \cite{zhao2017pyramid}. In our task, it can generate objects of different sizes at the same depth from the original objects. The size of the image patch ranges [0.64, 1.0] in Resized-Crop. Depth-Scaling rescales depth values by a random factor. It can set the same objects to different depth values. The random factor ranges [0.8, 1.2] in Depth-Scaling. Fig. \ref{fig: augmentation} illustrates our data augmentation.

\subsubsection{Loss functions} We adopt the loss function in \cite{wang2023g2}. This loss function ${L(z,z^*)}$ between output depth maps $z$ and ground-truth (GT) $z^*$ includes scale-adaptive loss ${L_{sa}}(z,z^*)$ and multi-scale scale-invariant gradient loss ${L_{sg}}(z,z^*)$. 

The loss function are defined as follows
\begin{equation}
\label{eq: g2loss}
\begin{aligned}
& {{L_{sa}}(z,z^*)} = \frac{1}{N}{\sum_{i=1}^N}|T_z-T_{z^*}| + \frac{1}{N_v+\epsilon}{\sum_{v=1}^{N_v}}|z_v-{z^*}_v|, \\
& {{L_{sg}}(z,z^*)} = \frac{1}{N}{\sum_{k=0}^3}{\sum_{i=1}^N}|\nabla_{hw}(\rho_k(T_z-T_{z^*}))|, \\
& {L(z,z^*)} ={{L_{sa}}(z,z^*)}+0.5{{L_{sg}}(z,z^*)},
\end{aligned}
\end{equation}
where $ T_z =(z-{\bar{z}}) / (\delta^{'}_z+\epsilon) $ and $ T_{z^*} =({z^*}-{\bar{z^*}}) / (\delta^{'}_{z^*}+\epsilon) $. $\delta^{'}_z$ and $\delta^{'}_{z^*}$ are the mean deviation of output depth maps $z$ and GT $z^*$. $\rho_k(.)$ is a down-sampling function on different resolutions $1/2^k$. $\nabla_{hw}$ is the gradient in $h$ and $w$ directions by the Sobel operator. $N$ and $N_v$ are the valid pixel numbers of GT and input sparse depth map, respectively. $\epsilon$ is a very small value to prevent a zero denominator.

\subsubsection{Implement details} Our network is initialized by Xavier Normal \cite{glorot2010understanding}. The optimizer is AdamW with a learning rate of 0.0002 and weight decay of 0.05. The batch size is 64 on two 3090 GPUs. We adopt cosine learning rate decay with 300 epochs. We use the Automatic Mixed Precision (AMP) of PyTorch to accelerate the training, which may slightly impair the final performance.

\begin{table*}[!h]
\caption{Comparison with released baseline models on sparse depth maps with randomly sampled 0.1\%/1\%/10\% valid pixels. The bold indicates the best result, and the underline indicates that our model achieves the second-best result. \label{tab:release_syn}}
\centering
\resizebox{\textwidth}{!}{
\begin{tabular}{l|cccccccccccc|c}
\hline
\multirow{2}{*}{Methods} & \multicolumn{2}{c}{Ibims} & \multicolumn{2}{c}{NYUv2} & \multicolumn{2}{c}{KITTI} & \multicolumn{2}{c}{DIODE} & \multicolumn{2}{c}{ETH3D} & \multicolumn{2}{c|}{Sintel} & \multirow{2}{*}{Rank $\downarrow$ } \\ 
\cline{2-13}
& Rel & RMSE & Rel & RMSE & Rel & RMSE & Rel & RMSE & Rel & RMSE & Rel & RMSE & \\ 
\hline
NLSPN{\textsuperscript{nyu}} \cite{park2020non} & 0.106 & 3.99 & 0.056 & 7.06  & 0.208 & 14.55 & 0.264 & 9.65  & 1.151 & 10.83 & 5.572  & 13.14 & 4.8 \\
NLSPN{\textsuperscript{kitti}} \cite{park2020non} & 0.098 & 4.64 & 0.095 & 11.27 & 0.240 & 15.25 & 0.333 & 11.19 & 1.168 & 12.01 & 8.038  & 18.04 & 7.1 \\
GuideNet \cite{tang2020learning} & 0.238 & 7.76 & 0.170 & 17.67 & 0.472 & 22.31 & 0.544 & 15.78 & 1.445 & 14.83 & 11.045 & 30.39 & 11.7 \\
TWISE \cite{imran2021depth} & 0.118 & 5.67 & 0.117 & 15.22 & 0.304 & 27.87 & 0.327 & 13.27 & 1.207 & 13.14 & 10.177 & 22.39 & 9.4 \\
MDANet \cite{ke2021mdanet} & 0.159 & 6.79 & 0.225 & 22.59 & 0.302 & 19.25 & 0.272 & 12.60 & 0.554 & 8.12  & 6.232  & 18.87 & 8.2 \\
EMDC \cite{hou2022learning} & 0.194 & 8.85 & 0.178 & 19.85 & 0.214 & 15.46 & 0.365 & 14.51 & 0.126 & 3.39  & 4.204  & 13.12 & 7.6 \\
SemAttNet \cite{nazir2022semattnet} & 0.184 & 8.83 & 0.315 & 31.48 & 0.325 & 30.67 & 0.313 & 17.96 & 0.928 & 10.10 & 6.781  & 25.34 & 10.6 \\
CFormer{\textsuperscript{nyu}} \cite{zhang2023completionformer} & 0.117 & 4.01 & 0.075 & 9.22  & 0.324 & 17.81 & 0.275 & 9.64  & 1.656 & 14.76 & 6.281  & 15.73 & 7.0 \\
CFormer{\textsuperscript{kitti}} \cite{zhang2023completionformer} & 0.206 & 7.44 & 0.093 & 10.66 & 0.533 & 22.09 & 0.504 & 14.14 & 2.426 & 22.40 & 18.661 & 27.86 & 10.7 \\
\textcolor{red}{LRRU} \cite{wang2023lrru} & 0.072 & 4.53 & 0.093 & 12.92 & 0.318 & 18.76 & 0.240 & 11.29 & 0.306 & 5.45 & 0.997 & 9.71 & 5.2\\
G2MD \cite{wang2023g2} & 0.018 & 1.70 & 0.027 & \textbf{4.99}  & 0.156 & 12.13 & 0.148 & 7.36  & 0.282 & 3.54  & 0.815  & 5.28  & 2.1 \\
\textcolor{red}{DFU} \cite{wang2024improving}  & 0.092 & 4.95 & 0.098 & 13.98 & 0.277 & 17.66 & 0.266 & 11.18 & 0.726 & 8.18 & 2.623 & 12.59 & 5.8\\
\rowcolor{gray!15} \textbf{Ours} & \textbf{0.012} & \textbf{1.51} & \textbf{0.025} & \underline{5.11}  & \textbf{0.065} & \textbf{8.00}  & \textbf{0.137} & \textbf{7.29}  & \textbf{0.051} & \textbf{1.98}  & \textbf{0.080}  & \textbf{4.24}  & \textbf{1.1} \\
\hline
\end{tabular}
}
\end{table*}

\begin{table*}[!t]
\caption{ Comparison with released baseline models on sparse depth maps with 4/8/16/32/64-line LiDAR points from KITTI and on raw depth maps with holes from NYUv2. Notably most baselines are trained on KITTI while it is unseen for our model.
\label{tab: release_real}}
\centering
\resizebox{\textwidth}{!}{
\begin{tabular}{l|cccccccccccc|c}
\hline
\multirow{3}{*}{Methods} & \multicolumn{2}{c}{\multirow{2}{*}{NYUv2}} & \multicolumn{10}{c|}{KITTI} & \multirow{3}{*}{Rank $\downarrow$ } \\ 
\cline{4-13}
& \multicolumn{2}{c}{~} & \multicolumn{2}{c}{4line} & \multicolumn{2}{c}{8line} & \multicolumn{2}{c}{16line} & \multicolumn{2}{c}{32line} & \multicolumn{2}{c|}{64line} & \\ 
& Rel & RMSE & Rel & RMSE & Rel & RMSE & Rel & RMSE & Rel & RMSE & Rel & RMSE & \\ 
\hline
NLSPN{\textsuperscript{nyu}} \cite{park2020non} & 0.025 & 5.49  & 0.193 & 16.64 & 0.077 & 10.54 & 0.046 & 7.67  & 0.038 & 6.33  & 0.035 & 5.61  & 6.9  \\
NLSPN{\textsuperscript{kitti}} \cite{park2020non} & 0.044 & 10.17 & 0.188 & 13.35 & 0.103 & 8.96  & 0.041 & 5.24  & 0.023 & 3.45  & 0.012 & 2.24  & 3.8  \\
GuideNet \cite{tang2020learning} & 1.580 & 97.28 & 0.567 & 28.18 & 0.275 & 16.87 & 0.115 & 8.76  & 0.054 & 5.26  & 0.022 & 3.17  & 10.7  \\
TWISE \cite{imran2021depth} & 0.084 & 10.97 & 0.372 & 41.35 & 0.134 & 21.27 & 0.066 & 16.32 & 0.034 & 10.39 & 0.019 & 6.55  & 9.6  \\
MDANet \cite{ke2021mdanet} & 0.298 & 24.58 & 0.208 & 16.08 & 0.130 & 10.34 & 0.051 & 5.47  & 0.026 & 3.49  & 0.013 & 2.23  & 6.3  \\
EMDC \cite{hou2022learning} & 0.729 & 63.82 & 0.485 & 24.77 & 0.367 & 22.94 & 0.267 & 20.96 & 0.215 & 19.73 & 0.187 & 18.90 & 12.5 \\
SemAttNet \cite{nazir2022semattnet} & 0.283 & 25.56 & 0.258 & 24.76 & 0.213 & 25.37 & 0.106 & 17.70 & 0.047 & 10.67 & 0.023 & 6.74  & 10.8  \\
CFormer{\textsuperscript{nyu}} \cite{zhang2023completionformer} & 0.335 & 26.57 & 0.225 & 17.52 & 0.105 & 12.23 & 0.049 & 8.06  & 0.039 & 6.38  & 0.036 & 5.76  & 9.4  \\
CFormer{\textsuperscript{kitti}} \cite{zhang2023completionformer} & 0.029 & 6.52  & 0.262 & 16.13 & 0.085 & 9.27  & 0.040 & 5.66  & \textbf{0.021} & 3.56  & \textbf{0.012} & 2.30  & 4.3  \\
\textcolor{red}{LRRU} \cite{wang2023lrru} & 0.059 & 13.68 & 0.226 & 16.57 & 0.107 & 10.45 & 0.042 & 5.84 & 0.022 & \textbf{3.29} & 0.013 & \textbf{2.20} & 5.0\\
G2MD \cite{wang2023g2} & \textbf{0.015} & \textbf{3.69}  & 0.070 & 9.88  & 0.045 & 7.25  & 0.035 & 5.88  & 0.028 & 4.76  & 0.025 & 4.23  & 4.2  \\
\textcolor{red}{DFU} \cite{wang2024improving}  & 0.098 & 17.47 & 0.176 & 14.93 & 0.089 & 9.72  & 0.042 & 5.96 & 0.022 & 3.43 & 0.013 & 2.21 & 4.8\\
\rowcolor{gray!15} \textbf{Ours} & \underline{0.019} & \underline{4.23}  & \textbf{0.055} & \textbf{8.21}  & \textbf{0.037} & \textbf{6.31}  & \textbf{0.029} & \textbf{5.21}  & 0.025 & 4.40  & 0.022 & 3.81  & \textbf{3.0} \\
\hline
\end{tabular}}
\end{table*}

\section{Experiments}
\label{section: experiments}
\subsection{Settings}
\subsubsection{Training data} Our network is trained on a mixed dataset, which is collected from four indoor/outdoor and real/synthetic datasets including Matterport3D \cite{chang2017matterport3d}, HRWSI \cite{xian2020structure}, vKITTI \cite{gaidon2016virtual}, and UnrealCV \cite{wang2023g2}. Sparse depth maps are generated by the data pipeline in \cite{wang2023g2}. The training dataset is augmented by Resized-Crop and Depth-Scaling introduced in Section \ref{section: implementation}. 

\subsubsection{Testing data} Our network is tested on six unseen datasets including Ibims \cite{koch2018evaluation}, KITTI \cite{geiger2012we}, NYUv2 \cite{silberman2012indoor}, DIODE \cite{vasiljevic2019diode}, ETH3D \cite{schops2017multi}, and Sintel \cite{butler2012naturalistic}. We comprehensively evaluate the performance of our models on different types of sparse depth maps. 

Firstly, sparse depth maps with 0.1\%/1\%/10\% valid pixels are randomly sampled from GT on all the six datasets following \cite{park2020non} \cite{zhang2023completionformer} \cite{tang2020learning} \cite{wang2023g2}. Secondly, 4/8/16/32/64-line LiDAR points are sampled from KITTI using the camera intrinsics following \cite{imran2021depth} \cite{zhang2023completionformer}. Thirdly, raw depth maps with holes are provided by NYUv2. Notably, the latter two types of sparse depth maps cannot be applied to the other four datasets. 

\subsubsection{Baselines} \textcolor{red}{Our model is compared with ten recent baselines including NLSPN \cite{park2020non}, GuideNet \cite{tang2020learning}, TWISE \cite{imran2021depth}, MDANet \cite{ke2021mdanet}, EMDC \cite{hou2022learning}, SemAttNet \cite{nazir2022semattnet}, CFormer \cite{zhang2023completionformer}, LRRU \cite{wang2023lrru}, G2MD \cite{wang2023g2}, and DFU \cite{wang2024improving}.} Codes of all these baselines are officially released by their authors. We compare our model to these baselines with both officially released models by the authors and retrained ones on our training data. 

Firstly, we directly use twelve \textit{officially released models} of these baselines to ensure their superior performance including NLSPN\textsuperscript{nyu} and CFormer\textsuperscript{nyu} trained on NYUv2 dataset; NLSPN\textsuperscript{kitti}, GuideNet, TWISE, MDANet, SemAttNet, CFormer\textsuperscript{kitti}, LRRU, and DFU trained on KITTI dataset; EMDC and G2MD trained on other datasets. 

Secondly, we \textit{retrain these baselines} on our training data to avoid the impact of different training data. \textcolor{red}{Notably, all data augmentations are also utilized during retraining for a fair comparison.} The six baselines NLSPN, MDANet, CFormer, LRRU, G2MD, and DFU with higher accuracy are retrained for easy implementation in the test, denoted NLSPN*, MDANet*, CFormer*, LRRU*, G2MD*, and DFU*. These retrained models are used to only evaluate the performance of our network architecture.

\subsubsection{Metrics} We adopt the common metrics absolute relative error (Rel) and root mean squared error (RMSE) to evaluate all models. Their formulations are as follows
\begin{equation*}
\begin{aligned}
    & \text{Rel}=\frac{1}{N} \sum_{i=1}^{N} {\frac{|z_i - z^*_i|}{z^*_i}},  \\
    & \text{RMSE}=\sqrt{\frac{1}{N} \sum_{i=1}^{N} (z_i - z^*_i)^2}.
\end{aligned}
\end{equation*}
We also use the average rank (denoted Rank) of these metrics to report the overall performance of all models across datasets and metrics \cite{ranftl2020towards} \cite{yin2022towards} \cite{wang2023g2}. 

\subsection{Comparison with baselines}
We compare our model to the baselines with both officially released models by the authors and retrained models on our training data.

\subsubsection{Released baseline models} We compare to the twelve baseline models that are officially released by their authors. Table \ref{tab:release_syn} shows the average results on sparse depth maps with 0.1\%/1\%/10\% valid pixels, which are randomly sampled from GT depth maps on the six unseen datasets. Our model achieves superior performance with the lowest Rank when compared to all these baseline models. Specifically, our model almost achieves the best results on all test datasets using Rel and RMSE. The effectiveness of our model in this test benefits from our SP-Norm, the design of basic block and network architecture as well as diverse training data.

Table \ref{tab: release_real} shows the evaluated results on sparse depth maps with 4/8/16/32/64-line LiDAR points from KITTI and on raw depth maps with holes captured by Structured-Light sensor from NYUv2. Our model also achieves superior performance with the lowest Rank compared to these baselines. Specifically, our model achieves the best results on 4/8/16-line points from KITTI and the second-best results on raw depth maps with holes from NYUv2. 

Notably, the baselines CFormer\textsuperscript{kitti}, NLSPN\textsuperscript{kitti}, LRRU, and DFU perform better in the scenarios of 32/64-line LiDAR points on KITTI in Table \ref{tab: release_real}. The reason lies that most baseline models are trained on the train set of KITTI (our test dataset), while all test datasets are unseen for our model in training.  

\begin{table*}[!h]
\caption{Comparison with retrained baseline models on sparse depth maps with randomly sampled 0.1\%/1\%/10\% valid pixels.\label{tab:retrain_syn}}
\centering
\resizebox{\textwidth}{!}{
\begin{tabular}{l|cccccccccccc|c}
\hline
\multirow{2}{*}{Methods} & \multicolumn{2}{c}{Ibims} & \multicolumn{2}{c}{NYUv2} & \multicolumn{2}{c}{KITTI} & \multicolumn{2}{c}{DIODE} & \multicolumn{2}{c}{ETH3D} & \multicolumn{2}{c|}{Sintel} & \multirow{2}{*}{Rank $\downarrow$ } \\ 
\cline{2-13}
& Rel & RMSE & Rel & RMSE & Rel & RMSE & Rel & RMSE & Rel & RMSE & Rel & RMSE & \\ 
\hline
NLPSN* & 0.035 & 2.04  & 0.037 & 5.30  & 0.112 & 10.68 & 0.148 & 7.53  & 0.197 & 2.92  & 53.000 & 55.58 & 4.3 \\
MDANet* & 1.105 & 27.34 & 0.506 & 32.96 & 2.253 & 63.14 & 1.205 & 29.36 & 4.888 & 50.53 & 43.713 & 53.85 & 6.8 \\
CFormer* & 0.019 & 1.83  & 0.029 & 5.39  & 0.106 & 11.26 & 0.137 & 7.49  & 0.075 & 2.33  & 35.392 & 33.81 & 3.3 \\
\textcolor{red}{LRRU*}  & 0.164 & 6.79 & 0.241 & 24.63 & 0.442 & 27.60 & 0.251 & 14.33 & 0.261 & 6.69 & 4.003 & 23.36 & 4.7\\
G2MD* & 0.015 & 1.67  & 0.025 & \textbf{4.89}  & 0.083 & 9.45  & \textbf{0.135} & \textbf{7.26}  & 0.089 & \textbf{1.82}  & 0.353  & 4.62  & 1.7 \\
\textcolor{red}{DFU*}  & 0.239 & 8.71 & 0.329 & 29.16 & 0.485 & 29.96 & 0.319 & 15.42 & 0.385 & 8.29 & 9.508 & 23.37 & 5.7\\
\rowcolor{gray!15} \textbf{Ours} & \textbf{0.012} & \textbf{1.51}  & \textbf{0.025} & \underline{5.11}  & \textbf{0.065} & \textbf{8.00}  & \underline{0.137} & \underline{7.29}  & \textbf{0.051} & \underline{1.98}  & \textbf{0.080}  & \textbf{4.24}  & \textbf{1.5} \\
\hline
\end{tabular}
}
\end{table*}

\begin{table*}[!h]
\caption{ Comparison with retrained baseline models on sparse depth maps with 4/8/16/32/64-line LiDAR points from KITTI and on raw depth maps with holes from NYUv2. \label{tab:retrain_real}}
\centering
\resizebox{\textwidth}{!}{
\begin{tabular}{l|cccccccccccc|c}
\hline
\multirow{3}{*}{Methods} & \multicolumn{2}{c}{\multirow{2}{*}{NYUv2}} & \multicolumn{10}{c|}{KITTI} & \multirow{3}{*}{Rank $\downarrow$ } \\ 
\cline{4-13}
& \multicolumn{2}{c}{~} & \multicolumn{2}{c}{4line} & \multicolumn{2}{c}{8line} & \multicolumn{2}{c}{16line} & \multicolumn{2}{c}{32line} & \multicolumn{2}{c|}{64line} & \\ 
& Rel & RMSE & Rel & RMSE & Rel & RMSE & Rel & RMSE & Rel & RMSE & Rel & RMSE & \\ 
\hline
NLPSN* & 0.017 & 3.76 & 0.089  & 10.67 & 0.050 & 7.22 & 0.040  & 5.74  & 0.038 & 4.74 & 0.034 & 4.19 & 4.0 \\
MDANet* & 0.111 & 13.83& 2.642 & 76.84& 1.092 & 43.73& 0.252 & 16.94& 0.093 & 8.27 & 0.045 & 4.75 & 6.8 \\
CFormer* & 0.014    & 3.74    & 0.086 & 11.46& 0.047 & 7.50 & 0.033 & 5.71 & 0.027 & 4.63 & 0.024 & 3.98    & 2.7 \\
\textcolor{red}{LRRU*} & 0.050 & 11.26 & 0.474 & 29.04 & 0.431 & 22.81 & 0.147 & 10.60 & 0.062 & 5.87 & 0.029 & 4.23 & 5.2\\
G2MD* & \textbf{0.014} & \textbf{3.67} & 0.080    & 10.01   & 0.045    & 7.03    & 0.030    & 5.51    & 0.026    & 4.62    & 0.023    & 4.08 & 2.0 \\
\textcolor{red}{DFU*}  & 0.105 & 13.73 & 0.540 & 33.56 & 0.353 & 21.49 & 0.116 & 10.01 & 0.074 & 6.26 & 0.058 & 4.90 & 5.8\\
\rowcolor{gray!15} \textbf{Ours} & 0.019 & 4.23 & \textbf{0.055} & \textbf{8.21} & \textbf{0.037} & \textbf{6.31} & \textbf{0.029} & \textbf{5.21} & \textbf{0.025} & \textbf{4.40} & \textbf{0.022} & \textbf{3.81} & \textbf{1.5} \\
\hline
\end{tabular}
}
\end{table*}

\subsubsection{Retrained baseline models} We further retrain the baselines on our training data with data augmentations to avoid the impact of different training data. These retrained models help reveal the performance of our network architecture only, without the impact of different training data. Notably, only the six baselines with higher accuracy in Table \ref{tab: release_real} are retrained for easy implementation including NLSPN \cite{park2020non}, MDANet \cite{ke2021mdanet}, CFormer \cite{zhang2023completionformer}, LRRU \cite{wang2023lrru}, G2MD \cite{wang2023g2}, and DFU \cite{wang2024improving}. 

Tables \ref{tab:retrain_syn} and \ref{tab:retrain_real} show that our model still achieves superior performance with the lowest Rank when compared to these baseline models. Specifically, our model almost achieves the best results on all types of sparse depth maps. The effectiveness of our model in this test only benefits from our SP-Norm as well as the design of basic block and network architecture.




\subsubsection{\textcolor{red}{Complexity analysis}} 

\textcolor{red}{The computational costs of our model and the baselines during inference are comprehensively evaluated on a 3090 GPU at 320$\times$320 resolution using the Pytorch platform. Table \ref{tab:costs} presents the results with the metrics: speed, model parameters, runtime per forward pass, FLOPs, and memory. Our tiny model achieves the best accuracy with the fastest speed of 126.6 image/s, the second-low FLOPs of 15.4G, and the third-low memory of 330MB, when compared to the baseline models. Our large model achieves the best accuracy while maintaining a competitive speed, FLOPs, and memory.}
The inference speed mainly benefits from the composition of our basic block and lightweight decoder. When directly replacing our modified basic block with the original basic block of ConvNeXt V2 \cite{woo2023convnext}, the inference speed of our model drops from 126.6 image/s to 114.1 image/s.

\textcolor{red}{Notably, our model is fully developed based on the Pytorch platform, without requiring any unique operators beyond its support. In addition, the used base model ConvNeXt V2 \cite{woo2023convnext} primarily utilizes depth-wise convolution layers and standard convolution layers. The efficiency of these operators has been verified in MobileNets \cite{howard2019searching} and EfficentNets \cite{tan2019efficientnet}. It ensures the easy application of our models in real scenarios.}

\begin{table}[!t]
\caption{\textcolor{red}{Complexity analysis. All models are evaluated on a 3090 GPU at 320$\times$320 resolution. \label{tab:costs}}}
\centering
\resizebox{\columnwidth}{!}{
\begin{tabular}{lcccccc}
\hline
   \multirow{2}{*}{Methods}    & Speed & Param.  & Time   & FLOPs & Memory & Accuracy \\
          & (image/s)  & (M)     & (ms)   & (G)   & (MB)   & (Rank)   \\
\hline
NLSPN \cite{park2020non}    & 44.1      & 26.2   & 22.7  & 162.1 & 3558 & 8.9       \\
GuideNet \cite{tang2020learning}  & 69.3      & 62.6   & 14.4  & 55.0  & 5634 & 14.7      \\
TWISE \cite{imran2021depth}    & \underline{115.9} & \textbf{1.5}    & \underline{8.6}   & 32.2  & 2160 & 12.4      \\
MDANet \cite{ke2021mdanet}    & 21.9      & \underline{3.0}    & 45.6  & 53.9  & 1896 & 11.2      \\
EMDC \cite{hou2022learning}     & 44.4      & 5.3    & 22.5  & \textbf{7.8}   & \textbf{300}  & 10.6      \\
SemAttNet \cite{nazir2022semattnet} & 10.5      & 361.0  & 95.7  & 208.0 & 2028 & 13.6      \\
CFormer \cite{zhang2023completionformer}   & 11.1      & 82.6   & 90.2  & 129.8 & 694  & 11.8      \\
LRRU \cite{wang2023lrru} & 33.1      & 20.8   & 30.2  & 215.8 & 582  & 8.2       \\
G2MD  \cite{wang2023g2}    & 88.0      & 18.2   & 11.4  & 23.7  & \underline{324}  & 4.8       \\
DFU \cite{wang2024improving}       & 23.9      & 25.5   & 41.9  & 194.1 & 812  & 8.8       \\
\rowcolor{gray!15}
{Ours{\textsuperscript{T}}}    & \textbf{126.6}     & 35.0   & \textbf{7.9}   & \underline{15.4}  & 330  & 3.8       \\
\rowcolor{gray!15}
{Ours{\textsuperscript{S}}}    & 77.6      & 59.3   & 12.9   & 25.1  & 420  & \underline{2.2}      \\
\rowcolor{gray!15}
{Ours{\textsuperscript{B}}}    & 76.7      & 105.0  & 13.0   & 44.3  & 600  & 2.4      \\
\rowcolor{gray!15}
{Ours{\textsuperscript{L}}}    & 60.2      & 235.5  & 16.6   & 99.2  & 1176 & \textbf{1.9}      \\
\hline
\end{tabular}
}
\end{table}

\subsubsection{Visual comparison} 
\begin{figure*}[!t]
\centering
\includegraphics[width=\textwidth]{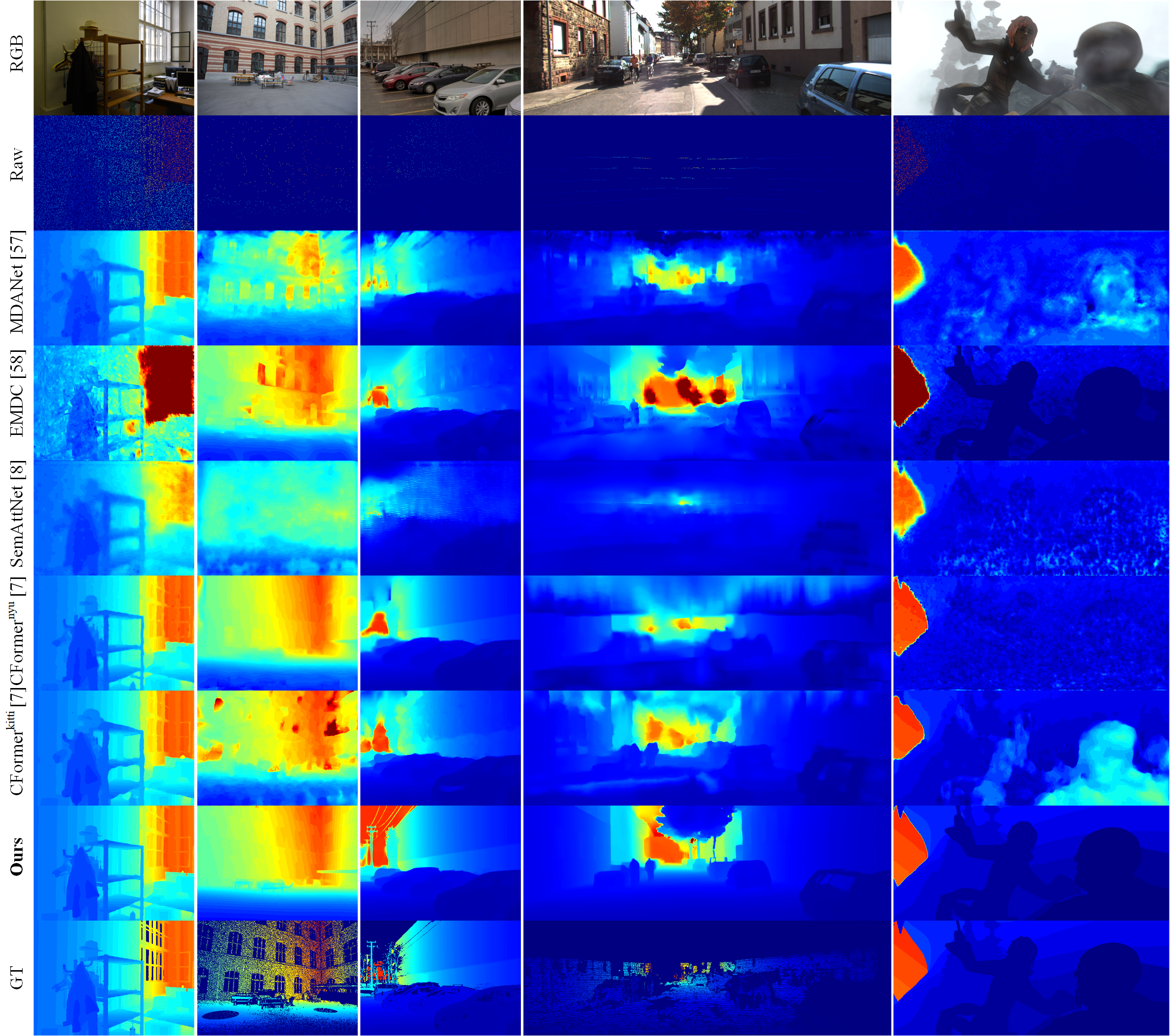}
\caption{Visual comparison on different scenes with various types of sparse depth maps from the experiments in Table \ref{tab:release_syn} and Table \ref{tab: release_real}. }
\label{fig: visual_3}
\end{figure*}

\begin{figure*}[!t]
\centering
\includegraphics[width=\textwidth]{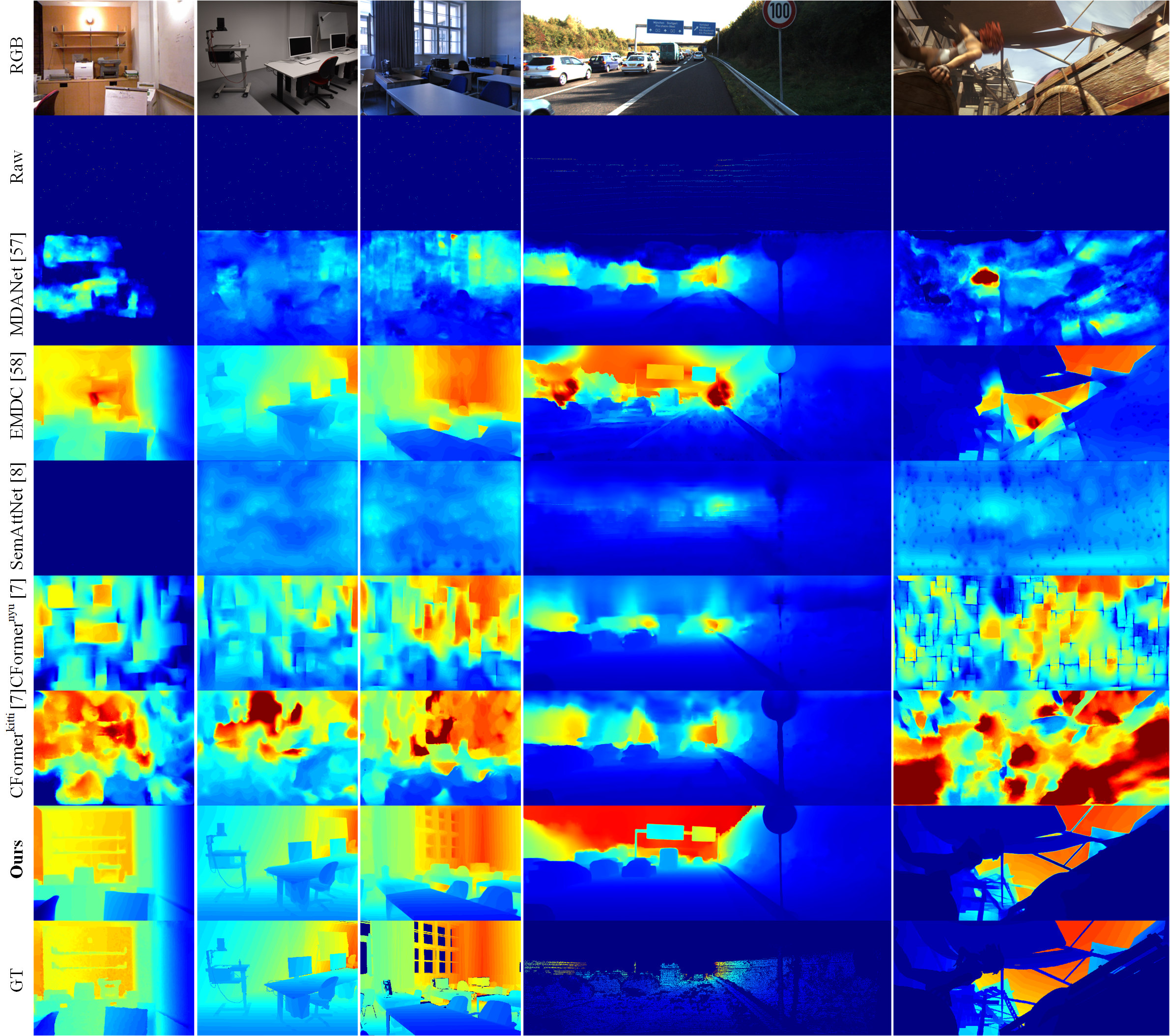}
\caption{Visual comparison on different scenes with various types of sparse depth maps from the experiments in Table \ref{tab:release_syn} and Table \ref{tab: release_real}. }
\label{fig: visual_5}
\end{figure*}

Fig. \ref{fig: visual_3} and Fig. \ref{fig: visual_5} show the visual results of our method and the baselines. All these visual results come from the experiments in Table \ref{tab:release_syn} and Table \ref{tab: release_real}. It is clear that our model always obtains high-quality depth maps in different scenes with various types of sparse depth maps. 

By comparison, the baseline models only work well in a few scenarios. In addition, our model achieves high-quality depth maps with clear structures and accurate scene scales compared with the baselines. It mainly benefits from the scale propagation property of our SP-Norm. 

\subsection{\textcolor{red}{Robustness evaluation}}
\label{section:robust}
\textcolor{red}{In this section, we verify the robustness of our models in two scenarios: raw depth maps with varying sparsity levels and dynamic environments with varying light conditions.}

\subsubsection{\textcolor{red}{Varying sparsity levels}} 
\begin{figure}[!t]
\centering
\includegraphics[width=\columnwidth]{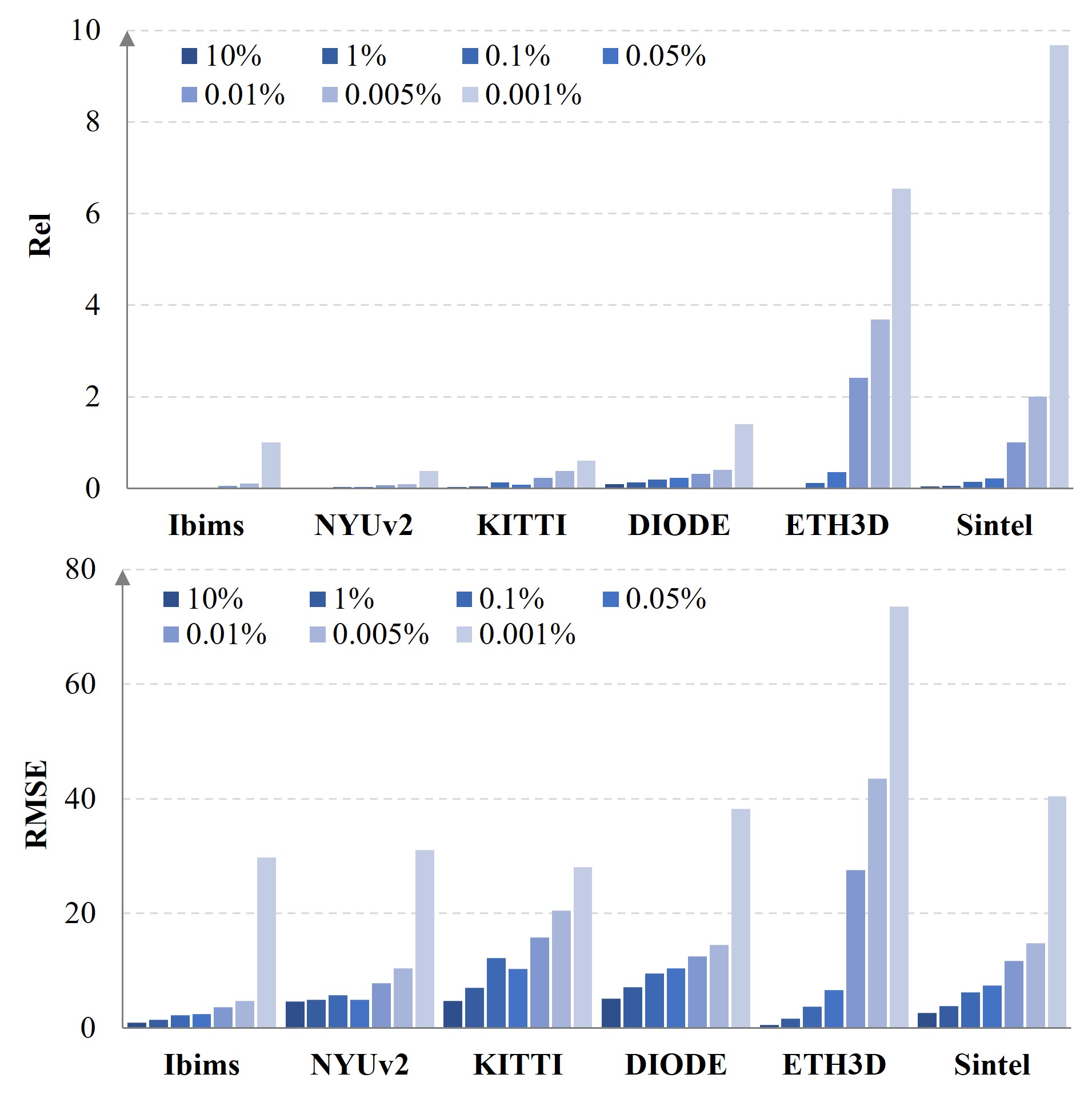}
\caption{\textcolor{red}{Performance in varying sparsity levels.}} 
\label{fig: sparsity}
\end{figure}

\textcolor{red}{We further evaluate our model in the scenarios with more varying sparsity levels 10\%, 1\%, 0.1\%, 0.05\%, 0.01\%, 0.005\%, and 0.001\% in Fig. \ref{fig: sparsity}. The results demonstrate that our model maintains effective performance even at the extremely low sparsity level 0.005\% in most scenarios. However, the performance degrades significantly at the sparsity level 0.001\%, which approximately corresponds to 2.05 valid pixels for 640$\times$320 resolution.} 

\textcolor{red}{It is mainly due to the limited scale information propagated from sparse depth points. Notably, the performance on ETH3D degrades rapidly when the sparsity level falls below 0.01\%. It is mainly because GTs in ETH3D are sparse. As a result, sparse depth maps in the input often contain fewer valid pixels after random sampling.}

\subsubsection{\textcolor{red}{Varying light conditions}} 
\begin{figure}[!t]
\centering
\includegraphics[width=\columnwidth]{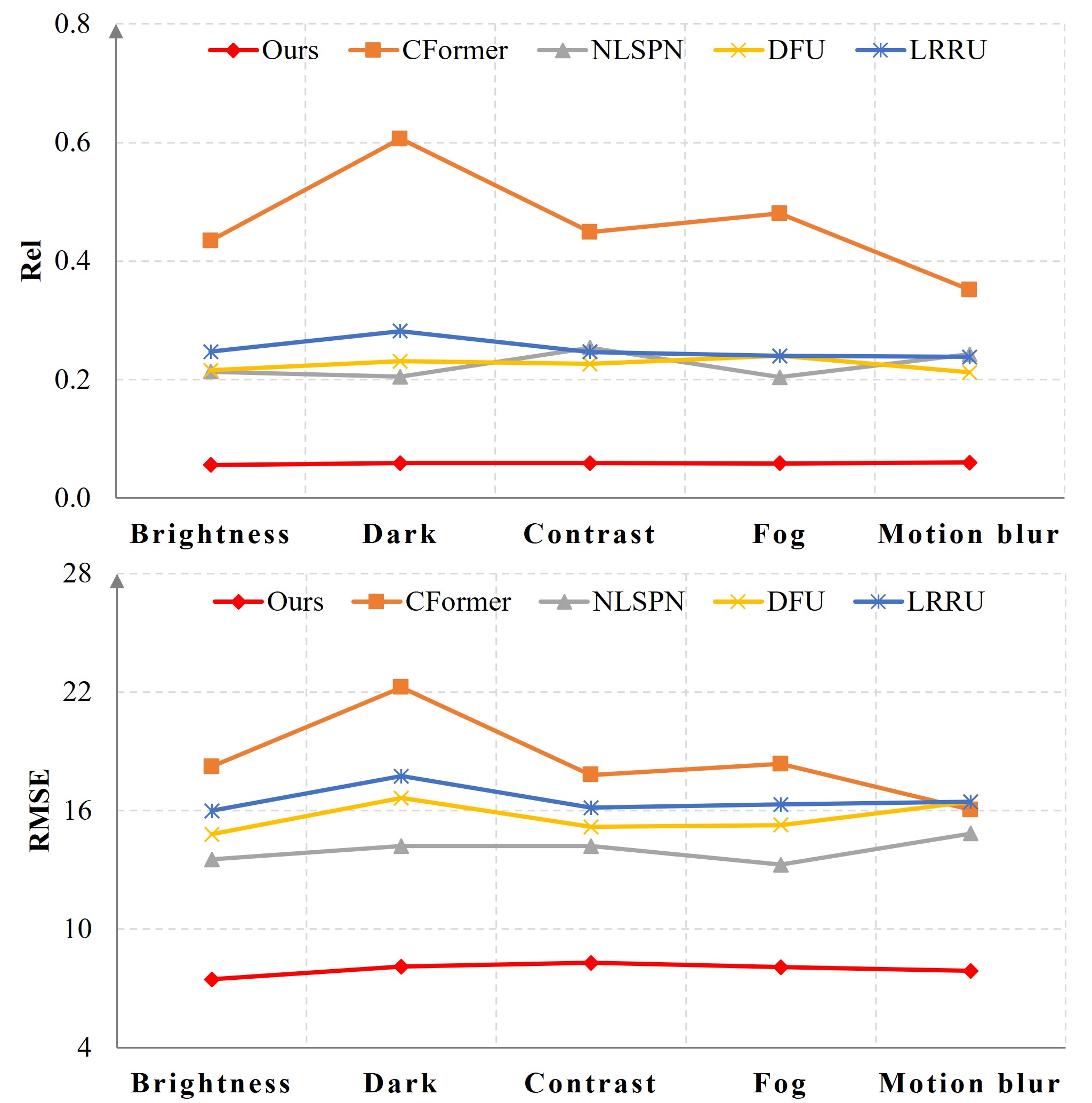}
\caption{\textcolor{red}{Performance in varying light conditions.}} 
\label{fig: dynamic}
\end{figure}

\textcolor{red}{We evaluate our model in dynamic environments in Fig. \ref{fig: dynamic}. The test data are from the public dataset KITTI-C \cite{NEURIPS2023_43119db5} using Eigen's test split, which contains a total of 18 conditions based on KITTI \cite{geiger2012we}. We select five conditions under varying light conditions including brightness, dark, contrast, fog, and motion blur.}

\textcolor{red}{The results show that our model stably achieves the best performance across varying light conditions compared to the released baseline models. We attribute the robustness of our model to the diverse training data, which covers various conditions in different environments.}

\subsection{Ablation studies}
\label{section:ablation}
In this section, we verify the effectiveness of each module in our method. The ablation studies are conducted on sparse depth maps with randomly sampled valid pixels on all test datasets. We train our models in this section with 90 epochs for fast implementation.

\begin{table}[!t]
\caption{ Ablation study of normalization strategies. \label{tab:ablation_norm}}
\centering
\resizebox{\columnwidth}{!}{
\begin{tabular}{lccccccc}
\hline
Methods & Ibims & NYUv2 & KITTI & DIODE & ETH3D & Sintel & {Rank $\downarrow$ } \\
\hline
\multicolumn{8}{c}{\textit{Rel}} \\
\hline
\multicolumn{1}{l|}{BN \cite{ioffe2015batch}}   & 0.051 & 0.073 & 0.111 & 0.187 & 0.330 & 2.270  & \multicolumn{1}{|c}{4.2} \\
\multicolumn{1}{l|}{IN \cite{ulyanov2016instance}}   & 0.095 & 0.048 & 0.139 & 0.199 & 0.495 & 2.741  & \multicolumn{1}{|c}{4.8} \\
\multicolumn{1}{l|}{LN \cite{ba2016layer}}   & 0.026 & 0.029 & 0.107 & 0.149 & 0.111 & 0.448  & \multicolumn{1}{|c}{2.8} \\
\multicolumn{1}{l|}{RZ \cite{bachlechner2021rezero}}   & 0.020 & 0.028 & 0.101 & 0.142 & 0.109 & 0.773  & \multicolumn{1}{|c}{2.2} \\
\rowcolor{gray!15} \multicolumn{1}{l|}{\textbf{Ours}} & \textbf{0.016} & \textbf{0.027} & \textbf{0.096} & \textbf{0.138} & \textbf{0.069} & \textbf{0.137}  & \multicolumn{1}{|c}{\textbf{1.0}} \\
\hline
\multicolumn{8}{c}{\textit{RMSE}} \\
\hline
\multicolumn{1}{l|}{BN \cite{ioffe2015batch}}   & 2.34 & 8.02 & 11.07 & 7.98 & 3.50 & 9.28  & \multicolumn{1}{|c}{4.2} \\
\multicolumn{1}{l|}{IN \cite{ulyanov2016instance}}   & 3.10 & 5.99 & 12.18 & 8.36 & 5.56 & 10.09  & \multicolumn{1}{|c}{4.8} \\
\multicolumn{1}{l|}{LN \cite{ba2016layer}}   & 1.88 & 5.14 & 9.37  & 7.59 & 2.21 & 5.30  & \multicolumn{1}{|c}{2.5} \\
\multicolumn{1}{l|}{RZ \cite{bachlechner2021rezero}}   & 1.85 & 5.16 & 10.05 & 7.50 & 2.18 & 5.44  & \multicolumn{1}{|c}{2.5} \\
\rowcolor{gray!15} \multicolumn{1}{l|}{\textbf{Ours}} & \textbf{1.70} & \textbf{5.10} & \textbf{9.37} & \textbf{7.34} & \textbf{1.73} & \textbf{4.72}  & \multicolumn{1}{|c}{\textbf{1.0}} \\
\hline
\end{tabular}}
\end{table}

\subsubsection{Different normalization strategies} SP-Norm is the key component of our network. We verify the effectiveness of our SP-Norm by replacing it with other normalization strategies in our network. It contains three conventional normalization layers including BN \cite{ioffe2015batch}, IN \cite{ulyanov2016instance}, LN \cite{ba2016layer}, and one non-normalization technique ReZero (RZ) \cite{bachlechner2021rezero}. 

Table \ref{tab:ablation_norm} shows the results of different normalization strategies with the metrics Rel and RMSE. Our SP-Norm significantly outperforms all these normalization strategies. Specifically, our SP-Norm achieves superior performance in all scenarios. It indicates that our SP-Norm is more suitable for generalizable depth completion, though these conventional normalization strategies achieve a great success in image analysis tasks. 

Notably, the non-normalization technique ReZero achieves the second-best results and outperforms all conventional normalization layers. It indicates that conventional normalization layers indeed hinder scale propagation in generalizable depth completion. It is in accordance with the analysis on the SP-property of conventional normalization layers in Section \ref{section: sp-norm}.

\subsubsection{\textcolor{red}{Apply our SP-Norm to other models}} \textcolor{red}{We further apply our SP-Norm component to other four models to show its effectiveness in Table \ref{tab:spnorm_baselines}. It is achieved by directly replacing normalization layers with our SP-Norm in these models. All other settings are kept the same to only evaluate our SP-Norm component. The results indicate that our SP-Norm component can effectively improve the generalization of these models.}

\textcolor{red}{Notably, we only modify the basic blocks in these models, which consist of convolution, normalization, and activation layers, to avoid the disturbance of their specially designed modules and pre-trained backbones. We also adopt a learning rate of 0.0001 to ensure the convergence of all models.}

\begin{table*}[!t]
\caption{\textcolor{red}{Apply our SP-Norm component to other models.} \label{tab:spnorm_baselines}}
\centering
\resizebox{\textwidth}{!}{
\begin{tabular}{l@{\hskip 5pt}c|cccccccccccc|c}
\hline
\multirow{2}{*}{Methods} &\multirow{2}{*}{SP-Norm} & \multicolumn{2}{c}{Ibims} & \multicolumn{2}{c}{NYUv2} & \multicolumn{2}{c}{KITTI} & \multicolumn{2}{c}{DIODE} & \multicolumn{2}{c}{ETH3D} & \multicolumn{2}{c|}{Sintel} & \multirow{2}{*}{Rank $\downarrow$ }\\ 
\cline{3-14}
& & Rel & RMSE & Rel & RMSE & Rel & RMSE & Rel & RMSE & Rel & RMSE & Rel & RMSE & \\ 
\hline
\multirow{2}{*}{NLSPN} &   & 0.102 & 3.56 & 0.101 & 8.93 & 0.157 & 15.24 & 0.193 & 8.72 & 0.238 & 3.56 & \textbf{47.151} & \textbf{51.66} & 1.8 \\
    & \cellcolor{gray!15} $\checkmark$ & \cellcolor{gray!15} \textbf{0.078} & \cellcolor{gray!15} \textbf{2.98} & \cellcolor{gray!15} \textbf{0.080} & \cellcolor{gray!15} \textbf{7.62} & \cellcolor{gray!15} \textbf{0.136} & \cellcolor{gray!15} \textbf{13.56} & \cellcolor{gray!15} \textbf{0.175} & \cellcolor{gray!15} \textbf{8.23} & \cellcolor{gray!15} \textbf{0.202} & \cellcolor{gray!15} \textbf{3.26} & \cellcolor{gray!15} 49.778 & \cellcolor{gray!15} 53.52 & \cellcolor{gray!15} \textbf{1.2} \\
\hline
\multirow{2}{*}{CFormer} &  & 0.030 & 2.15 & 0.037 & \textbf{5.63} & 0.108 & \textbf{10.17} & 0.146 & 7.86 & 0.085 & 2.13 & \textbf{33.687} & \textbf{32.58} & 1.7 \\
       & \cellcolor{gray!15} $\checkmark$ & \cellcolor{gray!15} \textbf{0.023} & \cellcolor{gray!15} \textbf{1.97} & \cellcolor{gray!15} \textbf{0.035} & \cellcolor{gray!15} 5.66 & \cellcolor{gray!15} \textbf{0.104} & \cellcolor{gray!15} 11.12 & \cellcolor{gray!15} \textbf{0.139} & \cellcolor{gray!15} \textbf{7.66} & \cellcolor{gray!15} \textbf{0.068} & \cellcolor{gray!15} \textbf{1.94} & \cellcolor{gray!15} 37.247 & \cellcolor{gray!15} 35.78 & \cellcolor{gray!15} \textbf{1.3} \\
\hline
\multirow{2}{*}{LRRU} &  & 0.164 & 6.79 & 0.241 & 24.63 & 0.442 & 27.60 & 0.251 & 14.33 & \textbf{0.261} & 6.69 & 4.003 & 23.36 & 1.9 \\
       & \cellcolor{gray!15} $\checkmark$ & \cellcolor{gray!15} \textbf{0.095} & \cellcolor{gray!15} \textbf{5.16} & \cellcolor{gray!15} \textbf{0.160} & \cellcolor{gray!15} \textbf{19.23} & \cellcolor{gray!15} \textbf{0.270} & \cellcolor{gray!15} \textbf{23.32} & \cellcolor{gray!15} \textbf{0.198} & \cellcolor{gray!15} \textbf{12.00} & \cellcolor{gray!15} 0.297 & \cellcolor{gray!15} \textbf{5.39} & \cellcolor{gray!15} \textbf{2.655} & \cellcolor{gray!15} \textbf{22.98} & \cellcolor{gray!15} \textbf{1.1} \\
\hline
\multirow{2}{*}{DFU} &  & 0.239 & 8.71 & 0.329 & 29.16 & 0.485 & 29.96 & 0.319 & 15.42 & 0.385 & 8.29 & \textbf{9.508}  & \textbf{23.37} & 1.8 \\
       & \cellcolor{gray!15} $\checkmark$ & \cellcolor{gray!15} \textbf{0.061} & \cellcolor{gray!15} \textbf{3.13} & \cellcolor{gray!15} \textbf{0.072} & \cellcolor{gray!15} \textbf{8.31}  & \cellcolor{gray!15} \textbf{0.148} & \cellcolor{gray!15} \textbf{13.84} & \cellcolor{gray!15} \textbf{0.180} & \cellcolor{gray!15} \textbf{8.66}  & \cellcolor{gray!15} \textbf{0.225} & \cellcolor{gray!15} \textbf{3.61} & \cellcolor{gray!15} 43.966 & \cellcolor{gray!15} 54.86 & \cellcolor{gray!15} \textbf{1.2} \\
\hline
\end{tabular}
}
\end{table*}

\subsubsection{Modifications in basic block} Our basic block is modified from the one of ConvNeXt V2 \cite{woo2023convnext} in three aspects. 

\textcolor{red}{Firstly, the most important modification is to fully replace LN with our SP-Norm. This modification is verified in the second row of the results in Table \ref{tab:ablation_block}. The basic block with our SP-Norm effectively improves the performance of our model compared to the one without SP-Norm in the third row.} 

Secondly, the GRN is removed from the basic block, which is a core operator in the ConvNeXt V2. This modification is verified in the second row of Table \ref{tab:ablation_block}. It indicates that the basic block without GRN improves the performance of our model compared to the basic block with GRN in the first row. 
The reason lies that the GRN will suppress features with lower global intensities. However, sparse depth maps in our task generally contain a large number of invalid pixels, which are represented by zero intensities. Therefore, the GRN may suppress features from sparse depth maps, which leads to inaccurate depth prediction. 

Thirdly, the activation function GELU is replaced with the RELU for faster inference. This modification is used to accelerate the inference speed of our network together with the other two modifications. We verify that these modifications improve the inference speed of our network from 114.1 image/s (original basic block) to 126.6 image/s (our basic block) on a 3090 GPU for the ``Tiny" model.

\begin{table}[!h]
\caption{ Ablation study of GRN, data augmentation, \textcolor{red}{and SP-Norm}. \label{tab:ablation_block}}
\centering
\resizebox{\columnwidth}{!}{
\begin{tabular}{c@{\hskip 5pt}c@{\hskip 5pt}cccccccc}
\hline
GRN & DA & \textcolor{red}{SP-Norm} & Ibims & NYUv2 & KITTI & DIODE & ETH3D & Sintel & {Rank $\downarrow$ } \\
\hline
\multicolumn{10}{c}{\textit{Rel}} \\
\hline
\checkmark & \checkmark & \multicolumn{1}{c|}{\checkmark}  & 0.042 & 0.034 & 0.098 & 0.184 & 0.198 & 0.541 & \multicolumn{1}{|c}{3.0} \\
\rowcolor{gray!15}       & \checkmark & \multicolumn{1}{c|}{\checkmark}  & \textbf{0.016} & \textbf{0.027} & \textbf{0.096} & \textbf{0.138} & \textbf{0.069} & \textbf{0.137} 
 & \multicolumn{1}{|c}{\textbf{1.0}} \\
 & \checkmark & \multicolumn{1}{c|}{}  & 0.026 & 0.029 & 0.107 & 0.149 & 0.111 & 0.448 & \multicolumn{1}{|c}{2.1} \\
 &        & \multicolumn{1}{c|}{\checkmark}  & 0.039 & 0.034 & 0.141 & 0.193 & 0.206 & 0.554 & \multicolumn{1}{|c}{3.9} \\
\hline
\multicolumn{10}{c}{\textit{RMSE}} \\
\hline
\checkmark & \checkmark & \multicolumn{1}{c|}{\checkmark}  & 2.55 & 4.65 & \textbf{8.63} & 8.55 & 3.47 & 6.03 & \multicolumn{1}{|c}{2.7} \\
\rowcolor{gray!15}       & \checkmark & \multicolumn{1}{c|}{\checkmark}  & \textbf{1.70} & 5.10 & 9.37 & \textbf{7.34} & \textbf{1.73} & \textbf{4.72} 
 & \multicolumn{1}{|c}{\textbf{1.4}} \\
 & \checkmark & \multicolumn{1}{c|}{}  & 1.88 & 5.14 & 9.37 & 7.59 & 2.21 & 5.30 & \multicolumn{1}{|c}{2.4} \\
 &        & \multicolumn{1}{c|}{\checkmark}  & 2.50 & \textbf{4.55} & 9.94 & 8.83 & 3.84 & 6.23 & \multicolumn{1}{|c}{3.4} \\
\hline
\end{tabular}
}
\end{table}

\subsubsection{Components of SP-Norm} Our SP-Norm comprises three components including the normalization operator, the SLP, and the multiplier. Table \ref{tab:ablation_variants} verifies the importance of each individual component. 

In the first row of these results, we remove the normalization operator (denoted Norm.) from our SP-Norm. This modification leads to unstable training and thereby results in convergence failure during training. It indicates the normalization operator is important for training our networks. 

In the second row, we replace the SLP with affine factors, which are widely used in conventional normalization layers such as BN \cite{ioffe2015batch}, IN \cite{ulyanov2016instance}, LN \cite{ba2016layer}. The modified network can still converge during training, nonetheless, the performance clearly degrades due to this modification. 

In the third row, we replace the multiplier (denoted Mul.) with the adder, which is widely used in residual learning. This also leads to the failure of convergence during training.

\begin{table}[!t]
\caption{ Ablation study of SP-Norm components. {``\textbackslash{}"} indicates that the networks do not converge. \label{tab:ablation_variants}}
\centering
\resizebox{\columnwidth}{!}{
\begin{tabular}{c@{\hskip 5pt}c@{\hskip 5pt}cccccccc}
\hline
Norm. & SLP & Mul. & Ibims & NYUv2 & KITTI & DIODE & ETH3D & Sintel & {Rank $\downarrow$ } \\
\hline
\multicolumn{10}{c}{\textit{Rel}} \\
\hline
     & \checkmark   & \multicolumn{1}{c|}{\checkmark}   & \textbackslash{}    &    \textbackslash{}   &    \textbackslash{}   &    \textbackslash{}   &    \textbackslash{}   &     \textbackslash{}   &    \multicolumn{1}{|c}{\textbackslash{}}  \\
\checkmark    &     & \multicolumn{1}{c|}{\checkmark}    & 0.018 & 0.027 & 0.111 & 0.141 & 0.081 & 0.281    & \multicolumn{1}{|c}{2.0}  \\
\checkmark    & \checkmark  & \multicolumn{1}{c|}{} &   \textbackslash{}    &     \textbackslash{}  &   \textbackslash{}    &    \textbackslash{}   &   \textbackslash{}    &    \textbackslash{}    &  \multicolumn{1}{|c}{ \textbackslash{}}   \\
\rowcolor{gray!15} \checkmark    & \checkmark  & \multicolumn{1}{c|}{\checkmark}  & \textbf{0.016} & \textbf{0.027} & \textbf{0.096} & \textbf{0.138} & \textbf{0.069} & \textbf{0.137} & \multicolumn{1}{|c}{\textbf{1.0}} \\
\hline
\multicolumn{10}{c}{\textit{RMSE}} \\
\hline
     & \checkmark   & \multicolumn{1}{c|}{\checkmark}   & \textbackslash{}    &    \textbackslash{}   &    \textbackslash{}   &    \textbackslash{}   &    \textbackslash{}   &     \textbackslash{}   &    \multicolumn{1}{|c}{\textbackslash{}}  \\
\checkmark    &     & \multicolumn{1}{c|}{\checkmark}    & 1.80 & \textbf{5.09} & 10.14 & 7.39 & 1.85 & 5.06    & \multicolumn{1}{|c}{1.8}  \\
\checkmark    & \checkmark  & \multicolumn{1}{c|}{} &   \textbackslash{}    &     \textbackslash{}  &   \textbackslash{}    &    \textbackslash{}   &   \textbackslash{}    &    \textbackslash{}    &  \multicolumn{1}{|c}{ \textbackslash{}}   \\
\rowcolor{gray!15} \checkmark    & \checkmark  & \multicolumn{1}{c|}{\checkmark}  & \textbf{1.70} & 5.10 & \textbf{9.37}  & \textbf{7.34} & \textbf{1.73} & \textbf{4.72} & \multicolumn{1}{|c}{\textbf{1.2}} \\
\hline
\end{tabular}}
\end{table}

\subsubsection{Data augmentation} Data augmentation in Section \ref{section: implementation} is utilized to improve the scale diversity of our training dataset. Table \ref{tab:ablation_block} verifies the effectiveness of the data augmentation (denoted DA). In the fourth row of these results, we train a model without data augmentation. In the second row, we train our final model with data augmentation. 

We can observe that the model without data augmentation is worse than our final model with data augmentation. It indicates that the data augmentation well improves the diversity of scene scales on the training dataset.

\subsubsection{Network Scalability} We provide the four versions of our networks in Table \ref{tab:config}. The models become larger from ``Tiny" to ``Large". We show the results of these models in Table \ref{tab:ablation_scalability}. The performance of our network can be consistently improved with larger models. It verifies that our network can be trained stably even using deeper models with more parameters. It also indicates that our network has the scalability to achieve better performance by adopting larger models. 

\begin{table}[!ht]
\caption{ Ablation study of model scalability among configurations of “Tiny”, “Small”, “Base”, and “Large”. \label{tab:ablation_scalability}}
\centering
\resizebox{\columnwidth}{!}{
\begin{tabular}{cccccccc}
\hline
       & Ibims & NYUv2 & KITTI & DIODE & ETH3D & Sintel & {Rank $\downarrow$ } \\
\hline
\multicolumn{8}{c}{\textit{Rel}} \\
\hline
\multicolumn{1}{c|}{Ours{\textsuperscript{T}}} & 0.014 & 0.026 & 0.074 & 0.139 & 0.074 & 0.089    & \multicolumn{1}{|c}{4.0} \\
\multicolumn{1}{c|}{Ours{\textsuperscript{S}}} & 0.013 & 0.025 & \textbf{0.062}& 0.139 & 0.055 & \textbf{0.078}& \multicolumn{1}{|c}{2.0} \\
\multicolumn{1}{c|}{Ours{\textsuperscript{B}}} & 0.013 & 0.026 & 0.071 & 0.138 & 0.067 & 0.079    & \multicolumn{1}{|c}{2.5} \\
\multicolumn{1}{c|}{Ours{\textsuperscript{L}}} & \textbf{0.012}& \textbf{0.025}& 0.065 & \textbf{0.137}& \textbf{0.051}& 0.080   & \multicolumn{1}{|c}{\textbf{1.5}} \\
\hline
\multicolumn{8}{c}{\textit{RMSE}} \\
\hline
\multicolumn{1}{c|}{Ours{\textsuperscript{T}}} & 1.61 & \textbf{5.09}& 8.58 & 7.35 & 2.07 & 4.36  & \multicolumn{1}{|c}{3.5} \\
\multicolumn{1}{c|}{Ours{\textsuperscript{S}}} & 1.58 & 5.14 & \textbf{7.86}& 7.29 & \textbf{1.77}& 4.19 & \multicolumn{1}{|c}{2.2} \\
\multicolumn{1}{c|}{Ours{\textsuperscript{B}}} & 1.53 & 5.15 & 8.28 & \textbf{7.26}& 1.95 & \textbf{4.15}& \multicolumn{1}{|c}{2.2} \\
\multicolumn{1}{c|}{Ours{\textsuperscript{L}}} & \textbf{1.51}& 5.11 & 8.00 & 7.29 & 1.98 & 4.24  & \multicolumn{1}{|c}{\textbf{2.2}} \\
\hline
\end{tabular}}
\end{table}

\section{Conclusion}
In this paper, we analyzed that a key design bottleneck of current deep neural networks resides in the conventional normalization layers, which limits the generalization of depth completion across different scenes. We proposed a novel scale propagation normalization method, SP-Norm, to enable scale propagation from input to output, and simultaneously preserve the normalization operator for easy convergence. We also explored a new network architecture based on the SP-Norm and the powerful ConvNeXt V2 for generalizable depth completion. Our network consistently achieves superior performance with efficient inference on unseen datasets with various types of sparse depth data compared to recent baselines. 

In our future work, we will explore a more robust and powerful model for generalizable depth completion through the utilization of pseudo labels, pre-training techniques, and unlabeled data. \textcolor{red}{In addition, the proportions between scales of the input and output in our SP-Norm are constant during testing for simplicity and efficiency. Inspired by the dynamic weights of the self-attention layer in Transformers, replacing constant proportions with dynamic ones may potentially improve the performance.}

{\appendix[Derivation Details]
Given two independent variables $p$ and $q$, we have
\begin{equation*}
\label{eq: base}
\begin{aligned}
        & E(p+q) = E(p) + E(q), \\
        & E(pq)  = E(p) E(q), \\
        & D(p+q) =D(p)+D(q), \\
        & D(pq)  =D(p)D(q)+E(p)^2 D(q)+E(q)^2D(p).
\end{aligned}
\end{equation*}
These equations will be frequently used in the below analyses.

\textit{Derivation of Eqn. (\ref{eq: spfortrdnorm}):} The formulation of conventional normalization is expressed in Eqn. (\ref{eq: norm})  and Eqn. (\ref{eq: affine}).  Based on these equations, we can get the mean and variance of the normalized data $\hat{d}_i$  as $E(\hat{d}_i)=0$ and $D(\hat{d}_i)=1$, respectively. 

Following Xavier Normal \cite{glorot2010understanding} and He Normal \cite{he2015delving}, we consider the affine factors $\alpha_i$, $\beta_i$, and normalized data $\hat{d}_i$ are independent with each other. By taking the mean and variance of Eqn. (\ref{eq: affine}), we have
\begin{equation}
\label{eq: mean_trd}
\begin{aligned}
E(z^{cd}_i) = E(\alpha_i \hat{d}_i + \beta_i) = E(\alpha_i)E(\hat{d}_i)+E(\beta_i) = E(\beta_i),
\end{aligned}
\end{equation}
\begin{equation}
\label{eq: var_trd}
\begin{aligned}
& D(z^{cd}_i) = D(\alpha_i \hat{d}_i + \beta_i) = D(\alpha_i\hat{d}_i)+D(\beta_i) \\
& = D(\alpha_i)D(\hat{d}_i)+E(\alpha_i)^2 D(\hat{d}_i) + E(\hat{d}_i)^2 D(\alpha_i) \\
& + D(\beta_i) = D(\alpha_i) + E(\hat{d}_i)^2 +D(\beta_i).
\end{aligned}
\end{equation}

Eqn. (\ref{eq: spfortrdnorm}) is derived from the Eqn. (\ref{eq: mean_trd}) and Eqn. (\ref{eq: var_trd}).

\textit{Derivation of Eqn. (\ref{eq: sppforspnorm}):}  We first denote $({\sum_{j=1}^n} w_{ij} {\hat{d}_j} + {b_i})$ in Eqn. (\ref{eq: spnorm}) as $f_i$ for convenience. We also consider that the parameters of the SLP $w_{ij}$, $b_i$, and normalized data $\hat{d}_i$ are independent with each other. We then have
\begin{equation}
\label{eq: mean_f}
\begin{aligned}
& E(f_i) = E({\sum_{j=1}^n} w_{ij} {\hat{d}_j} + {b_i}) = {\sum_{j=1}^n} E(w_{ij}\hat{d}_j)+E(b_i) \\
& = {\sum_{j=1}^n} E(w_{ij})E(\hat{d}_j)+E(b_i) = E(b_i), \\
\end{aligned}
\end{equation}
\begin{equation}
\label{eq: var_f}
\begin{aligned}
& D(f_i) = D({\sum_{j=1}^n} w_{ij} {\hat{d}_j} + {b_i}) = {\sum_{j=1}^n} D(w_{ij}\hat{d}_j)+D(b_i) \\
& = {\sum_{j=1}^n} (D(w_{ij})D(\hat{d}_j)+ E(w_{ij})^2 D(\hat{d}_j) + E(\hat{d}_j)^2 D(w_{ij})) \\ & + D(b_i) = n(D(w_{ij})+E(w_{ij})^2)+D(b_i).
\end{aligned}
\end{equation}

Because $f_i$ only depends on variables $w_{ij}$ and $b_i$ in Eqn. (\ref{eq: mean_f}) and Eqn. (\ref{eq: var_f}), we consider that $f_i$ is also independent to the input data $d_i$.  By taking the mean and variance of Eqn. (\ref{eq: spnorm}), we have
\begin{equation}
\label{eq: mean_spnorm}
\begin{aligned}
E(z^{sp}_i) = E(f_i d_i) = E(f_i)E(d_i) = E(b_i)E(d_i),
\end{aligned}
\end{equation}
\begin{equation}
\label{eq: var_spnorm}
\begin{aligned}
& D(z^{sp}_i) = D(f_i d_i) \\
& = D(f_i)D(d_i) + E(f_i)^2 D(d_i) + E(d_i)^2 D(f_i) \\
& = D(d_i)(n(D(w_{ij})+E(w_{ij})^2) +D(b_i) + E(b_i)^2) \\
& +E(d_i)^2 (n(D(w_{ij})+E(w_{ij})^2)+D(b_i)) \\
& = D(d_i) ({\Lambda} + {E(b_i)^2}) + {E(d_i)^2}{\Lambda },
\end{aligned}
\end{equation}
where ${\Lambda}=n(D(w_{ij}) + E(w_{ij})^2) + D(b_i)$. 

Eqn. (\ref{eq: sppforspnorm}) is derived from Eqn. (\ref{eq: mean_spnorm}) and Eqn. (\ref{eq: var_spnorm}).

\textit{Derivation of Eqn. (\ref{eq: initinput}):} \textcolor{red}{We consider that $d_i$ is the output of the first convolution layer of our network.} It can be expressed as $d_i={\sum_{j=1}^{n^0}}(w^0_{ij}d^0_j+b^0_i)$, where $d^0_j$ and $n^0$ are input data and input dimension of this convolution layer, respectively. $w^0_{ij}$ and $b^0_i$ are parameters of this convolution layer. Because the network is initialized by Xavier Normal \cite{glorot2010understanding}, we have $E(w^0_{ij})=0$, $D(w^0_{ij})=2/(n^0 + n^1)$, $E(b^0_i)=0$, and $D(b^0_i)=0$, where $n^1$ is the output dimensions of this convolution layer. 

We also consider that the parameters of the convolution $w^0_{ij}$, $b^0_i$, and the input data $d^0_j$ are independent with each other. By taking the mean and variance of the output data $d_i$ of this layer, we have
\begin{equation}
\label{eq: mean_input}
\begin{aligned}
& E(d_i) = E({{\sum_{j=1}^{n^0}}(w^0_{ij}d^0_j+b^0_i)}) = {\sum_{j=1}^{n^0}}E(w^0_{ij}d^0_j) \\ 
& = {\sum_{j=1}^{n^0}}E(w^0_{ij})E(d^0_j)=0,
\end{aligned}
\end{equation}
\begin{equation}
\label{eq: var_input}
\begin{aligned}
& D(d_i) = D({{\sum_{j=1}^{n^0}}(w^0_{ij}d^0_j+b^0_i)}) = {\sum_{j=1}^{n^0}}D(w^0_{ij}d^0_j) \\ 
& = {\sum_{j=1}^{n^0}}(D(w^0_{ij})D(d^0_j)+E(w^0_{ij})^2 D(d^0_j) + E(d^0_j)^2 D(w^0_{ij})) \\
& = {\sum_{j=1}^{n^0}} (\frac{2}{n^0 + n^1}(D(d^0_j)+E(d^0_j)^2)) \\
& = \frac{2n^0}{n^0 + n^1}(D(d^0_j)+E(d^0_j)^2)).
\end{aligned}
\end{equation}

Eqn. (\ref{eq: initinput}) is derived from Eqn. (\ref{eq: mean_input}) and Eqn. (\ref{eq: var_input}).

\bibliography{ref}

\vspace{11pt}

\bf{}\vspace{-33pt}
\begin{IEEEbiography}[{\includegraphics[width=1in,height=1.25in,clip,keepaspectratio]{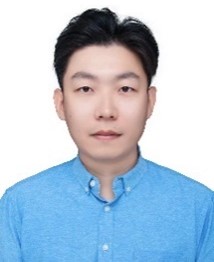}}]{Haotian Wang}
received the B.S. degree in electronic and information engineering in North China Electric Power University in 2017. He is currently pursuing the Ph.D. degree with the Institute of Artificial Intelligence and Robotics, Xi’an Jiaotong University. His research interests include 3D vision and multi-modal vision.
\end{IEEEbiography}

\vspace{11pt}

\bf{}\vspace{-33pt}
\begin{IEEEbiography}[{\includegraphics[width=1in,height=1.25in,clip,keepaspectratio]{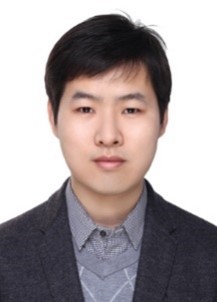}}]{Meng Yang}
(Member, IEEE) received the B.S. degree in information engineering and the Ph.D. degree in control science and engineering, Xi’an Jiaotong University, China, in 2008 and 2014, respectively. He was a visiting scholar with the University of California at San Diego, USA, from 2011 to 2012. He is currently an Associate Professor with the Institute of Artificial Intelligence and Robotics, Xi’an Jiaotong University. His research interests include machine vision and autonomous vehicle.
\end{IEEEbiography}

\vspace{11pt}

\bf{}\vspace{-33pt}
\begin{IEEEbiography}[{\includegraphics[width=1in,height=1.25in,clip,keepaspectratio]{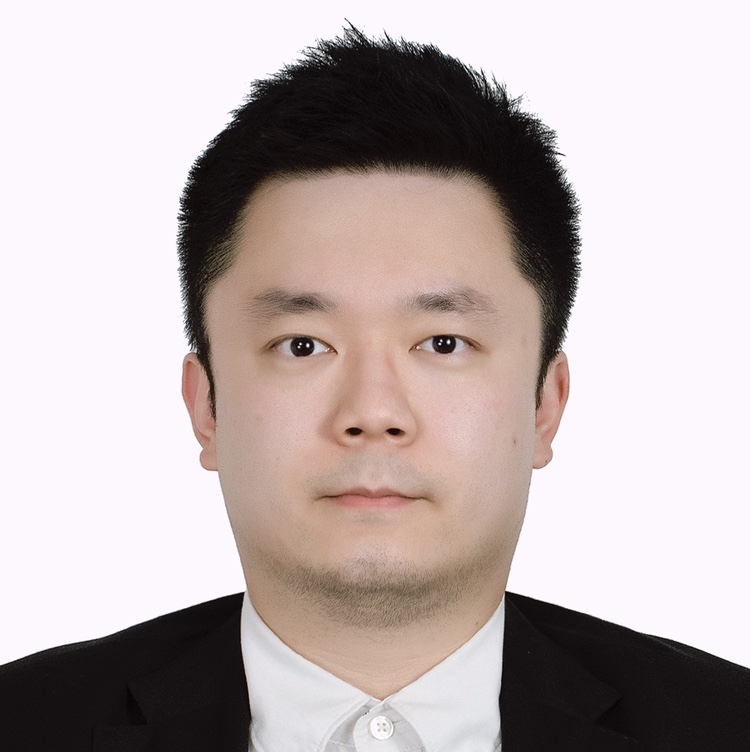}}]{Xinhu Zheng}
(Member, IEEE) received the Ph.D. degree in electrical and computer engineering from the University of Minnesota, Minneapolis, in 2022. He is currently an Assistant Professor with the Hong Kong University of Science and Technology (GZ). His current research interests are data mining in power systems, intelligent transportation system by exploiting different modality of data, leveraging optimization, and machine learning techniques. He currently serves as an Associated Editor for IEEE Transactions on Intelligent Vehicles.
\end{IEEEbiography}

\vspace{11pt}

\bf{}\vspace{-33pt}
\begin{IEEEbiography}[{\includegraphics[width=1in,height=1.25in,clip,keepaspectratio]{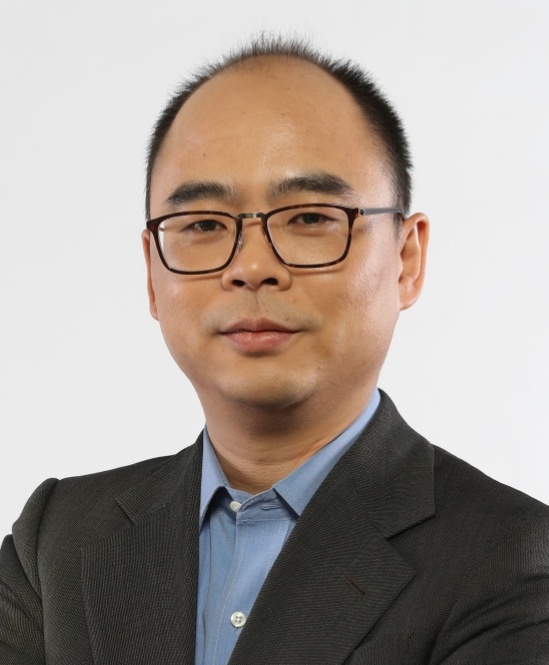}}]{Gang Hua}
(Fellow, IEEE) received the B.S. and M.S. degrees in automatic control engineering from Xi'an Jiaotong University, Xi'an, China, in 1999 and 2002, respectively, and the Ph.D. degree in electrical engineering and computer science with Northwestern University, Evanston, IL, USA, in 2006. He is currently the Vice President of Multimodal Experiences Lab in Dolby Laboratories Inc. He is an Associate Editor for TPAMI and MVA. He is the General Chair of ICCV’2025, the Program Chair of CVPR'2019\&2022, and the Area Chair of CVPR'2015\&2017 and ICCV'2011\&2017. He is the author of more than 200 peer reviewed publications in prestigious international journals and conferences. He holds 19 U.S. patents and has 15 more U.S. patents pending. He was the recipient of the 2015 IAPR Young Biometrics Investigator Award. He is an IAPR Fellow and an ACM Distinguished Scientist.

\end{IEEEbiography}

\vfill

\end{document}